\documentclass[10pt,twocolumn,letterpaper]{article}

\usepackage{cvpr}      
\usepackage[dvipsnames]{xcolor}

\definecolor{cvprblue}{rgb}{0.21,0.49,0.74}
\usepackage[pagebackref,breaklinks,colorlinks,citecolor=cvprblue]{hyperref}
\usepackage{multirow}
\def\paperID{10340} 
\def\confName{CVPR}
\def\confYear{2024}

\title{HOIDiffusion: Generating Realistic 3D Hand-Object Interaction Data}

\author{Mengqi Zhang$^{1*}$ \quad Yang Fu$^{1*}$ \quad Zheng Ding$^1$ \quad Sifei Liu$^2$ \\
\quad Zhuowen Tu$^1$ \quad Xiaolong Wang$^1$\\
$^1$UC San Diego \quad $^2$ NVIDIA}

\begin{document}

\twocolumn[{%
\maketitle

\centering
\vspace{-2.5em}
\includegraphics[width=\textwidth]{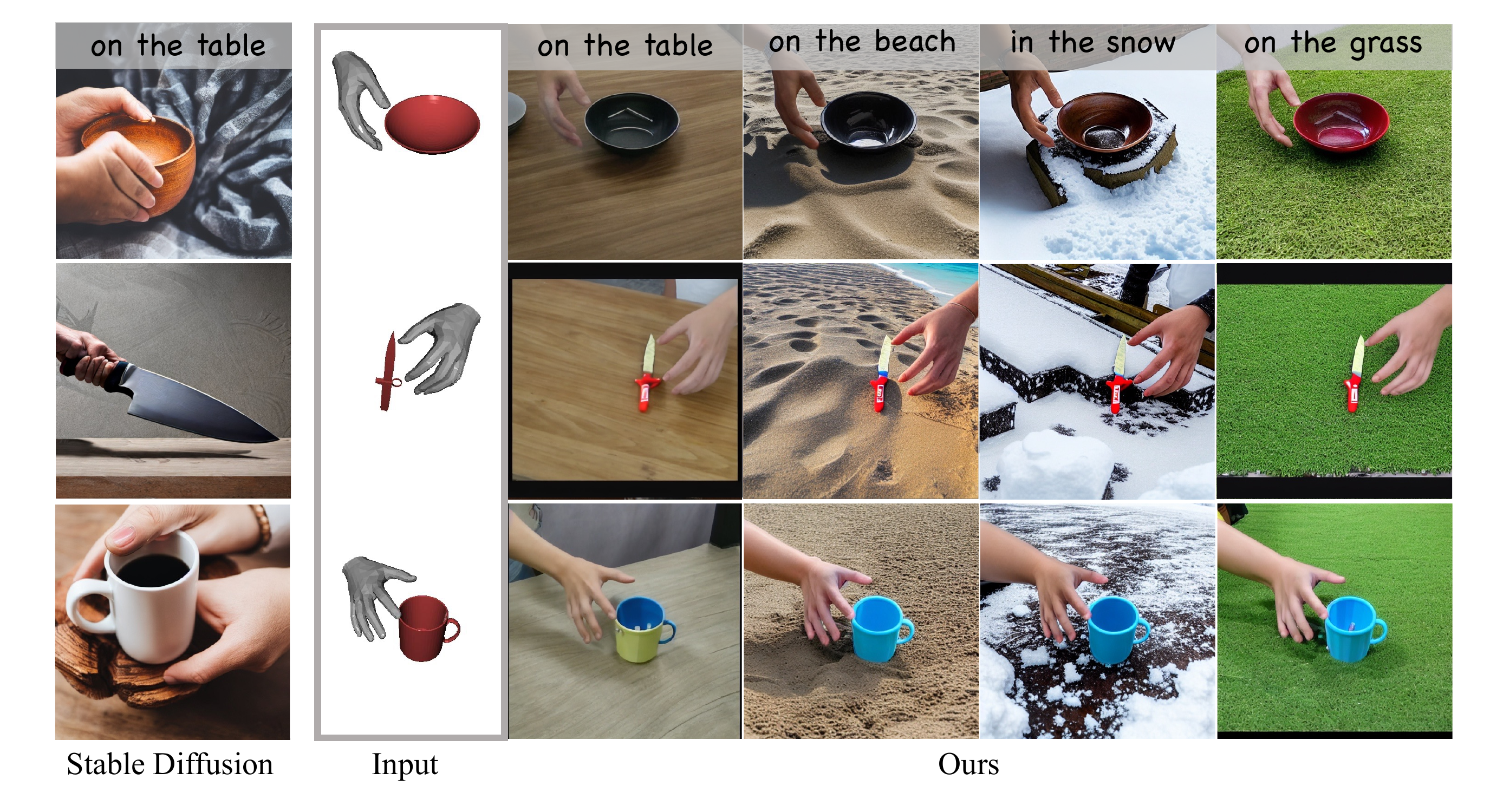} 
\vspace{-0.4in}
\captionof{figure}{\small (i) \textbf{Left}: Hand-object synthesis with Stable Diffusion model; (ii) \textbf{Right}: HOIDiffusion generates high-quality hand-object interaction images conditioned on physical structures and detailed text description. The model disentangles the geometry from appearance, exhibiting high generation diversity. Each row: We can fix the structure and control the style based on text inputs; Each column: We can fix the style and control the structure based on 3D structural inputs.}
\label{fig:teaser}
\vspace{1em}
 }]
\def\thefootnote{*}\footnotetext{Equal Contribution.}
\begin{abstract}
\vspace{-0.17in}
3D hand-object interaction data is scarce due to the hardware constraints in scaling up the data collection process. In this paper, we propose HOIDiffusion for generating realistic and diverse 3D hand-object interaction data. Our model is a conditional diffusion model that takes both the 3D hand-object geometric structure and text description as inputs for image synthesis. This offers a more controllable and realistic synthesis as we can specify the structure and style inputs in a disentangled manner. HOIDiffusion is trained by leveraging a diffusion model pre-trained on large-scale natural images and a few 3D human demonstrations. Beyond controllable image synthesis, we adopt the generated 3D data for learning 6D object pose estimation and show its effectiveness in improving perception systems. Project page: \href{https://mq-zhang1.github.io/HOIDiffusion}{https://mq-zhang1.github.io/HOIDiffusion}.
\end{abstract}
 
\vspace{-2em}
\section{Introduction}
\label{sec:intro}
Understanding how human hands interact with objects has been a long-standing problem in computer vision. Recently, researchers have tried to scale up such understandings by collecting videos on hand-object interactions~\cite{grauman2022ego4d,damen2018scaling}. Models trained with these datasets focus on performing hand-object relational reasoning in 2D space. To enable broader applications in robotics and VR/AR, more efforts have been spent on collecting hand-object interaction data with 3D annotations via multiple cameras~\cite{chao2021dexycb} and new labeling approaches with prepared object CAD models~\cite{hampali2020honnotate}. However, such a data collection process is not scalable and most datasets only contain dozens of objects. 

Given the recent advancement of generative modeling with diffusion process~\cite{ho2020denoising,song2020denoising}, can we leverage them to generate realistic 3D hand-object interaction data? While state-of-the-art diffusion models such as Dall-E2~\cite{ramesh2022hierarchical} and Stable Diffusion~\cite{rombach2022high} can generate realistic images given text instructions, they still fail quite often when it comes to capturing the details of how fingers are placed around the object. As shown on the left side of Figure~\ref{fig:teaser}, Stable Diffusion might not be able to output physically or geometrically plausible interactions and sometimes there are more than five fingers in a hand. Moreover, it is still unclear how to configure image outputs beyond text instructions and make the output correspond to 3D shapes and poses. 

In this paper, we propose to generate 3D hand-object interaction data, i.e., realistic images come with 3D ground-truths at the same time. Beyond realistic generation, we also enable controllable synthesis where the users can specify the geometry configuration and appearance in a disentangled manner. We achieve this by introducing a two-stage framework: We first synthesize the 3D geometric structure (shape and pose) of the hand and the object, and then we train a diffusion model conditioned on both the 3D structure and the text (indicating the style) to synthesize the corresponding RGB image. We visualize some synthesis results on the right side of Figure~\ref{fig:teaser}. In each row we generate images with the same 3D structural configuration but with different object and background styles; In each column, we fix the styles and synthesize images with different geometric structures. 

In the first stage of our framework, we generate a human grasp based on a given 3D object model. We apply the pre-trained GrabNet~\cite{taheri2020grab} for this task, which takes the object mesh as inputs for a Variational AutoEncoder and predicts different grasp poses as outputs. In the second stage, we train a diffusion model conditioning the hand-object geometric configurations. We fine-tune the pre-trained Stable Diffusion model~\cite{rombach2022high} with a few human demonstrations from the DexYCB dataset~\cite{chao2021dexycb}. We convert the hand-object geometry to estimated surface normals, segmentation, and hand keypoint 2D projection as conditional inputs for the new diffusion model, specifying the structure of the image to generate. The diffusion model will also take text inputs for specifying appearance. During training, we apply a background regularization strategy to reduce the bias brought by DexYCB which comes with the same clean background. The fine-tuned diffusion model leverages both the rich appearance information from the pre-trained model and the geometry information from the new conditional variables.

In our experiments, our method outperforms previous approaches on hand-object image synthesis with more physically plausible interactions. The disentangled design provides flexible control of geometry and appearance. Interestingly, the model shows a strong generalization ability to different text prompts when changing the foreground and background appearance. We use several metrics to evaluate generation fidelity to real datasets and visual alignment with provided prompts. The results show an improvement compared to baselines. Additionally, with a generated dataset with both images and corresponding 3D geometry using our pipeline, we can use it to train an object pose estimator as a downstream application. Our experiments indeed show such realistic synthesized data is very helpful in improving the perception metrics.

\begin{figure*}[t]
\centering
\scalebox{1.0}{
    \begin{tabular}{c}
\includegraphics[width=1\textwidth]{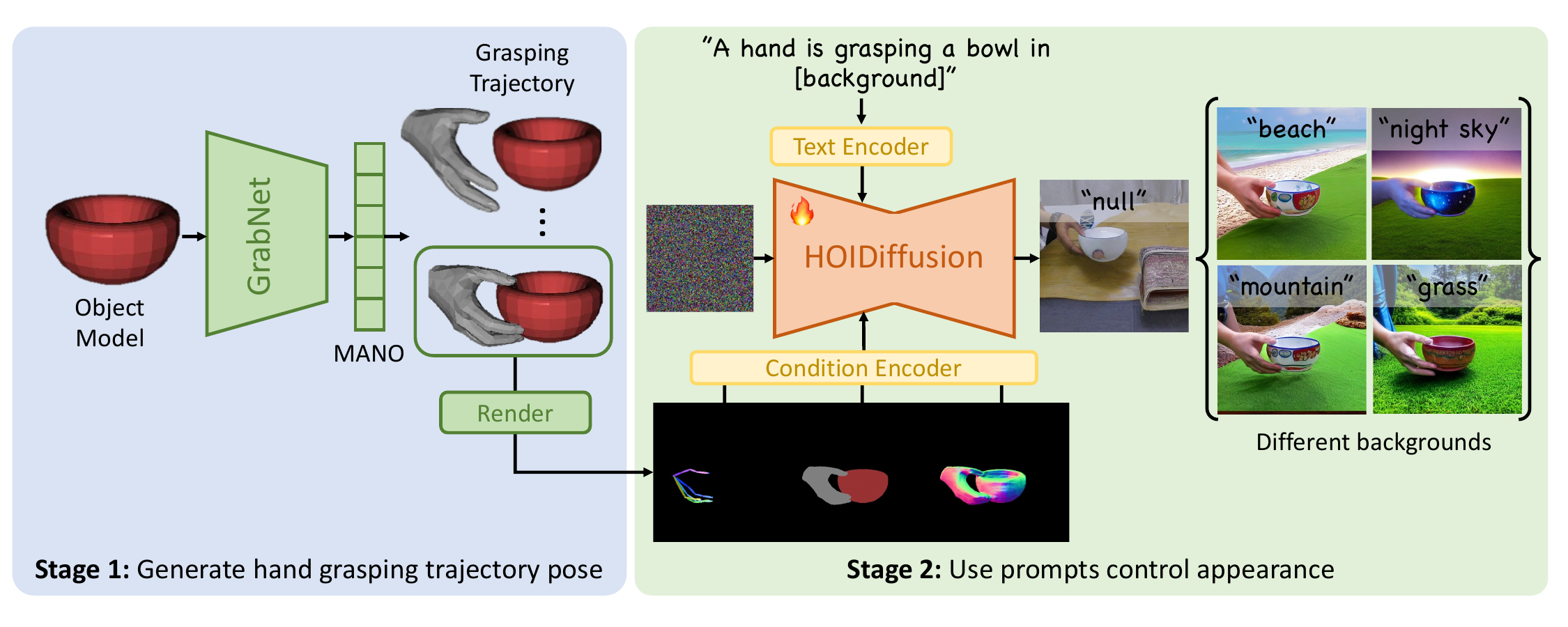}\\
\end{tabular}
}
\vspace{-0.2in}
\caption{\textbf{Pipeline.} We propose a two-stage pipeline to synthesize hand-object-interaction data. During the first stage, we utilize a pretrained GrabNet to output 3D hand poses given by a single object model. Then in the second stage, we use those 3D hand poses along with segmentation maps, normal maps and skeletons to conditionally generate high-quality HOI data.}
\vspace{-0.2in}
    \label{fig:pipeline}
\end{figure*}
\section{Related work}
\noindent \textbf{Hand-Object Interaction Dataset.} The understanding of hand-object interactions has been a long-standing problem~\cite{hamer2010object,oikonomidis2011full,ballan2012motion,panteleris20153d,sridhar2016real}. More recently, the data-driven approaches~\cite{hasson2019learning,oberweger2019generalized,doosti2020hope,hasson2020leveraging,liu2021semi,cao2021reconstructing} have shown significant advancement in hand-object shape reconstruction and pose estimation. At the heart of this progress, is the collection of hand-object interaction data. To obtain 3D annotations, existing datasets~\cite{hasson2019learning,taheri2020grab,fan2023arctic,ye2021h2o,hampali2020honnotate} collect videos with attached sensors or mocap markers to track hand pose or utilize optimized algorithms to facilitate annotations. 2D annotations are also provided with manual labeling~\cite{chao2021dexycb}. However, these approaches are time-consuming not scalable. Recently more data collection pipelines~\cite{grady2021contactopt,yang2022oakink,qin2022dexmv,liu2022hoi4d} have taken advantage of detection and segmentation techniques to automatically acquire annotations. However, even though this method eases the difficulty, estimations may not be precise enough, and manual annotations might still be required for accurate 3D annotations. The diversity of the data is also relatively small given the repetitive patterns in videos. In this paper, we propose a new effective data generation method facilitated by generative models for hand-object interaction images with full 3D annotations.

\noindent \textbf{Hand Grasp Generation.}  Hand grasp generation given an object model~\cite{corona2020ganhand, karunratanakul2020grasping, jiang2021hand, brahmbhatt2019contactgrasp, grady2021contactopt, zhang2021manipnet, zhu2021toward, christen2022d} is of vital importance in our method to provide an ending pose for grasping trajectory. Most approaches estimate or further refine the grasping hand pose by predicting the contact map between hands and objects~\cite{brahmbhatt2020contactpose, jiang2021hand,grady2021contactopt}. Other methods ~\cite{taheri2020grab, karunratanakul2020grasping,fan2023arctic} predict the MANO parameter hand representation introduced in ~\cite{romero2022embodied} with variational autoencoder or implicit function architectures. Additionally, The hand parameter prediction can be integrated as a component of the whole human body~\cite{taheri2022goal}. By taking advantage of these grasp predictors, we are able to obtain satisfying ending poses for hand trajectory generation.

  \noindent \textbf{Diffusion Models.}  Diffusion models~\cite{ho2020denoising,song2020denoising,rombach2022high} which learn to denoise images from Gaussian distributions, emerge recently and perform photo-realistic image synthesis with more stable training process compared to other generative models~\cite{kingma2013auto,goodfellow2014generative}. Many successive advancements occur in this field~\cite{nichol2021improved, dhariwal2021diffusion,nichol2021glide,ramesh2022hierarchical,saharia2022photorealistic}. More relevantly, special tokens are introduced~\cite{gal2022image,ruiz2023dreambooth} to fine-tune the model, enabling personalized text-to-image generation. With these amazing generation results, researchers~\cite{burg2023data,wu2023datasetdm} start investigating the possibility of utilizing diffusion models as a new source of data for classification or segmentation tasks. Besides using text inputs, conditional generation with a given layout or scratch achieves impressive performance after the appearance of CoAdapter~\cite{mou2023t2i} and ControlNet~\cite{zhang2023adding}. These works unveiled the large potential of diffusion models to control or edit images according to users' demands. However, these models still suffer from generating realistic hand-object-interaction images, such as not being object-agnostic, inaccurate geometry, and generating hands with missing fingers or unnatural poses. In our work, we focus on how to utilize physical conditions, including normal, hand skeleton projection and segmentation to construct a 3D-aware model for scalable hand-object-image generation with flexible geometry and appearance control.

\section{Method}

In order to generate scalable hand-object-interaction images, three expectations need to be satisfied: (i) The model should generate realistic images that are consistent with the geometric description of the specified object. (ii) It should retain the prompt-editing capabilities inherent in the stable diffusion model while incorporating controllable conditions. (iii) The model should have a better generalization ability to synthesize images of unseen instances or categories.

To meet the above requirements, we propose a two-stage approach. For the first stage, our goal is to establish the conditions, primarily the hand-grasping trajectory, for the subsequent stage. Specifically, we utilized a generalizable VAE model trained on large-scale 3D physical data to obtain the ending pose and interpolated the trajectory using spherical linear interpolation. For the second stage, we extracted multiple geometry structures either from the generated grasping trajectory or from images with natural hand-grasping actions through rendering and off-the-shelf estimators to finetune a controllable Stable Diffusion, which enables precise pose control at inference. Furthermore, a background regularization strategy is introduced in our pipeline to mitigate edit ability degradation brought by finetuning. The entire pipeline is shown in Figure \ref{fig:pipeline}.

\subsection{Preliminary}
Denoising diffusion model~\cite{ho2020denoising} is a kind of new generative model with competitive performance and more stable training. It consists of two main processes: diffusion and denoising. In the forward process, randomly sampled noises are added to original images, which can be mathematically simplified as  $q(x_t|x_0) = \mathcal{N}(x_t; \sqrt{\overline{\alpha}_t}x_0,(1-\overline{\alpha}_t)I)$, where ${\alpha}_t$ is hyperparameters control noise scheduling, and $\overline{\alpha}_t = \prod_{s=1}^t \alpha_s$. After $T$ steps, the distribution shifted from image space to approximate standard Gaussian distribution. And U-Net model is trained to predict the added noise. During inference, an image is initialized from the Gaussian distribution and the model removes the noise within $T$ steps, with each step $t$ formulated as  
\begin{equation}
\hat{\epsilon_t} = f_{\theta}(x_{t}, t)
\label{eq:forward1}
\end{equation}
where $f_{\theta}$ is the U-Net model. The estimated image at the next time step can be derived and written as
\begin{equation}
x_{t-1} = \frac{1}{\sqrt{\alpha_t}}(x_t-\frac{1-\alpha_t}{\sqrt{1-\overline{\alpha}_t}}\hat{\epsilon_t})+\sigma_tz
\label{eq:backward1}
\end{equation}
The simpler version of training loss is:$||\epsilon-f_{\theta}(\sqrt{\overline{\alpha}_t}x_0+\sqrt{1-\overline{\alpha}_t}\epsilon,t)||^2$. Therefore diffusion models are able to generate photorealistic images. 

\subsection{Hand Grasping Trajectory Generation}
In the first stage of our framework, given a randomly transformed object model, we require a hand-grasping trajectory to effectively reach the object. To this end, we adopted an interpolation method using GrabNet, a model for generating hand grasps conditioned on BPS (Body Part Segmentation)~\cite{prokudin2019efficient} derived from object models. Through extensive training on a large dataset, it has learned to accurately map contact between hands and objects, showcasing strong generalization capabilities for unseen scenarios. Utilizing GrabNet, we can determine the final grasp hand MANO parameters, which include joint pose and hand shape. Given the ending pose and the object's position, we can approximate an initial starting point—positioned vertically above the contact surface and within a certain distance—using zeroed MANO parameters. Subsequently, we employ spherical linear interpolation between this starting point and the ending pose to create a smooth hand-grasping trajectory. As shown in Figure \ref{fig:pipeline} Stage 1, the grasping parameters can be obtained along the trajectory and we could acquire different ground-truth annotations such as segmentation, mask, and depth maps through rendering.

\subsection{Hand-Object-Interaction image Synthesis}
Our Hand-Object-Interaction image synthesis module mainly consists of two components: structure control and appearance regularization, to disentangle the geometry and appearance separately.

\noindent\textbf{Structure Control}  
Given the hand-grasp trajectory, we extract multiple geometric conditions and leverage the advanced Stable Diffusion models to generate realistic images that are consistent with the given conditions. To precisely control the hand-object image generation, three structure conditions are distilled to control the generation. i) To synthesize realistic hands without missing fingers which is the problem encountered in original Stable Diffusion, hand position information is essentially required. Instead of directly using MANO parameter vectors as guidance, we projected the skeleton information onto the image space as visual control, which can be denoted as $s^i$. ii) Additionally, to mitigate the inter-disturbance of hand object areas and degradation in performance brought by occlusion, hand-object segmentation ($m^i$) is used to provide clear boundaries to separate areas, and coarse object shape priors. iii) Finally, we also apply an estimated normal map ($n^i$) to seize the surface geometry with lighting. The forward process defined in Equation \ref{eq:forward1} can now be defined as $\hat{\epsilon_t^i} = f_{\theta}(x^i_{t}, t, [s^i,m^i,n^i])$. With the above controls, the structure information could be disentangled from the appearance, and thus during the inference, we could seamlessly synthesize an accurate image with new poses.
\begin{figure*}[t]
\centering
\scalebox{0.95}{
    \begin{tabular}{c}
\includegraphics[width=1\textwidth]{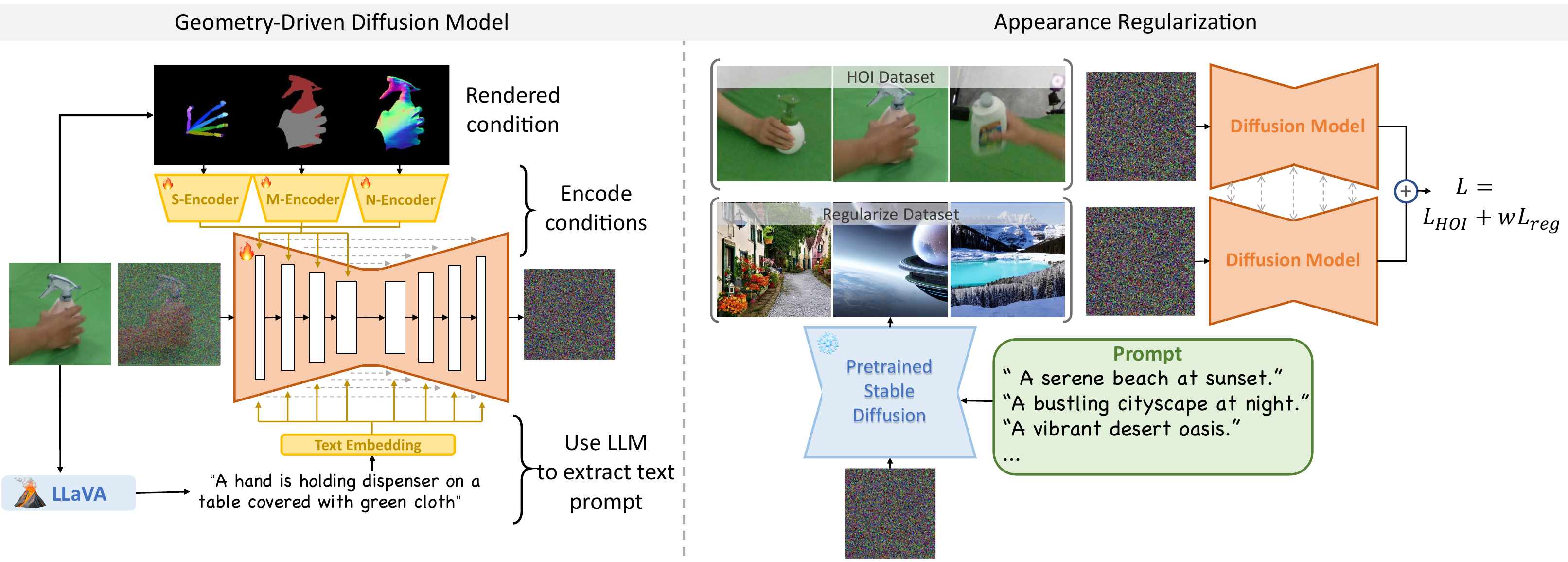}\\

\end{tabular}
}
\vspace{-5mm}
\caption{\textbf{Model Figure.} We inject three conditional encoders into the stable diffusion model. We utilize both the HOI datasets and high-quality background images to train HOIDiffusion. The background images are synthesized using the scenery prompts. The texts sent to the model are output by LLaVA for detailed description.}
    \label{fig:main_model}
    
\end{figure*}

\noindent\textbf{Appearance Regularization}  With the above component, we are capable of synthesizing images aligned with diverse condition geometries during inference. However, a notorious drawback of fine-tuning is its tendency to converge or overfit quickly to the training dataset's style, thereby significantly reducing image diversity. To mitigate this problem and fully harness the capabilities of text-to-image diffusion models for flexible style transformation via prompts, we introduce an appearance regularization method combined with classifier-free guidance ~\cite{ho2022classifier} as shown in Figure \ref{fig:main_model}. Specifically, in addition to using the original Hand-Object Interaction (HOI) training dataset, we synthesize batches of high-quality scenery images with the pre-trained text-to-image diffusion model. The prompts for these images are generated by the large language model ChatGPT, forming what we refer to as a ``background buffer". During training, we intermittently utilize these background images for regularization, ensuring it does not detrimentally impact performance. Meanwhile, the paired blank conditions are applied as classifier-free guidance, corresponding to the background region in HOI data. The objective becomes:
\begin{equation}
\begin{aligned}
\mathcal{L} = E_{x_0,x_r,\epsilon,\epsilon_r}[||\epsilon-f_{\theta}(\sqrt{\overline{\alpha}_t}x_0+\sqrt{1-\overline{\alpha}_t}\epsilon,t)||^2\\
+w_r||\epsilon_r-f_{\theta}(\sqrt{\overline{\alpha}_t}x_r+\sqrt{1-\overline{\alpha}_t}\epsilon_r,t)||^2]
\label{eq:loss}
\end{aligned}
\end{equation}
where $w_r$ is the regularization weight, which is set to 1 in our experiments. $x_r$, $\epsilon_r$ are input from the background buffer and corresponding added noise. In addition to the buffer, we also use the large multimodal model LLaVA~\cite{liu2023visual} to caption our training images with detailed descriptions of foreground appearance and background, forcing the model aware of diverse texts with the assistance of the CLIP~\cite{radford2021learning} text encoder.    
\section{Experiment}

\subsection{Implementation Details}
To achieve scalable synthesis and improve generalization ability, the training dataset is required to encompass diverse backgrounds and comprehensive annotations. For these purposes, the DexYCB dataset is selected for its high-quality images from varied viewpoints of hand gestures, with human demonstrations and diverse background settings. To prevent overfitting and preserve the text-driven editing ability of diffusion models, we adopt the learning rate of $10^{-5}$ during training. About 50,000 steps prove sufficient to achieve satisfactory generation quality. When utilizing images from background buffers, we provide vacant skeleton projection and mask conditions, corresponding to the background regions in HOI training data images. This design also can be viewed as a classifier-free guidance that drives the model aware of the condition control effects. The entire training process costs approximately 12 hours on eight A100 GPUs. 

\subsection{Evaluation and Baseline Comparison}
\textbf{Baselines}  We compare the results with four baseline models customized for our setting: (1) a fine-tuned LDM ~\cite{rombach2022high} without condition modules, working as an unconditional generation task; (2) a modified DreamBooth~\cite{ruiz2023dreambooth} fine-tuned model, with proposed specific token agnostically applied to all object categories, complemented by regularization data from our trained conditional stable diffusion; (3) Affordance Diffusion~\cite{ye2023affordance}; (4) ControlNet with same multiple condition input. It is important to mention that the Affordance Diffusion model is only used for comparison on contact recall evaluation, with quantitative results reported from the original paper.

\noindent \textbf{Image Synthesis Quality}  To assess the generation quality, we adopt commonly used metrics FID~\cite{heusel2017gans} to evaluate the fidelity of our generated images to real datasets. Additionally, we also present the comparison results with baselines on the Inception Score~\cite{salimans2016improved} and sFID~\cite{nash2021generating}. Our reference batch consists of 50k randomly selected images from the training dataset in total. For the sample batch, we synthesize hand-grasping images corresponding to all randomly rotated object models, some of which are unseen during training. The total sample size is 5k. We refer readers to Table \ref{tab:baseline} for a comprehensive overview of the comparison.

\noindent \textbf{Appearance Alignment}  Furthermore, to demonstrate the flexible control over object and background appearance, we use another metrics CLIPScore~\cite{hessel2021clipscore} originally designed for image captioning to automatically evaluate the alignment level between generated images and corresponding prompts. We randomly sample 50 instances from all object models with fixed structure conditions. Multiple appearance descriptions generated by ChatGPT are applied to each instance. The results are shown in Table \ref{tab:baseline}. Notably, our method demonstrates superior overall quality of generated images, with higher fidelity to real data and improved alignment with appearance-controlling texts.

\begin{table}[t]
    \vspace{-2mm}
    \centering\small
    
    \scalebox{0.9}{
    \begin{tabular}{l c c c c}
         \toprule
         Method & FID$\downarrow$ & sFID$\downarrow$ & IS$\uparrow$ &CLIPScore $\uparrow$\\
         \midrule
         LDM~\cite{rombach2022high}  & 63.71 & 119.10 & 6.81 & 0.68  \\
         DreamBooth~\cite{ruiz2023dreambooth} & 134.40 & 92.82 & \textbf{7.99} & 0.75  \\
         ControlNet~\cite{zhang2023adding} & 87.99 & 248.56 & 6.60 & 0.77 \\
         \midrule
          HOIDiffusion (Ours) & \textbf{55.22} & \textbf{91.28} & 7.73 & \textbf{0.78}\\
    
         \bottomrule
    \end{tabular}
    }
    \vspace{-2mm}
    \caption{\small \textbf{Quantitative comparison with previous baseline methods}. All models are trained on the DexYCB. We use FID to directly measure the synthesis quality of generated hand-object interaction images.
    sFID is a recently proposed metric to evaluate image quality using higher-level spatial features. IS is measured for diversity and CLIPScore is to evaluate generated images alignment with provided prompts.
    }
    \vspace{-0.2in}
    \label{tab:baseline}
\end{table}

\begin{figure*}
\centering
\scalebox{0.98}{
    \setlength\tabcolsep{0.3pt}
    \renewcommand{\arraystretch}{0.2}
    \begin{tabular}{cccccccc}
         \includegraphics[width=0.125\textwidth]{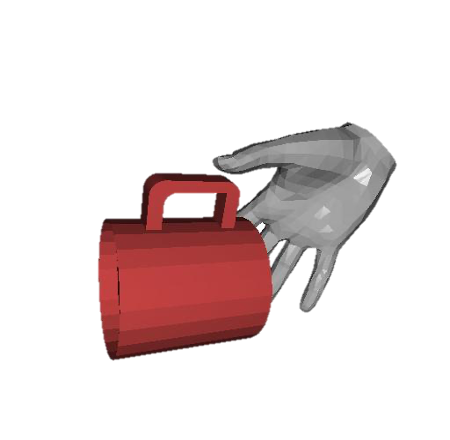} &
         \includegraphics[width=0.125\textwidth]{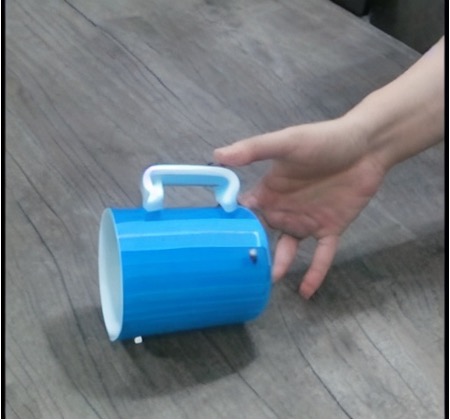}  &
         \includegraphics[width=0.125\textwidth]{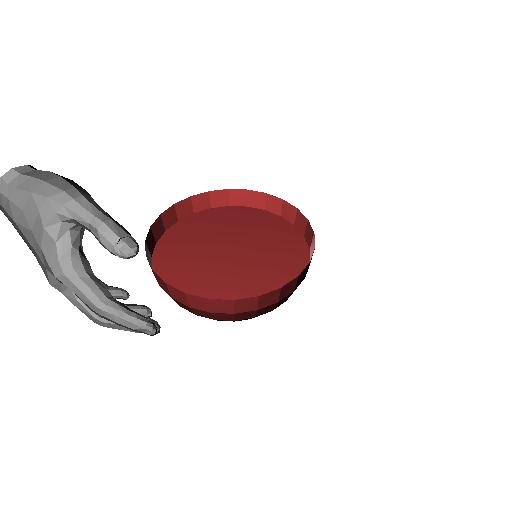}  &
         \includegraphics[width=0.125\textwidth]{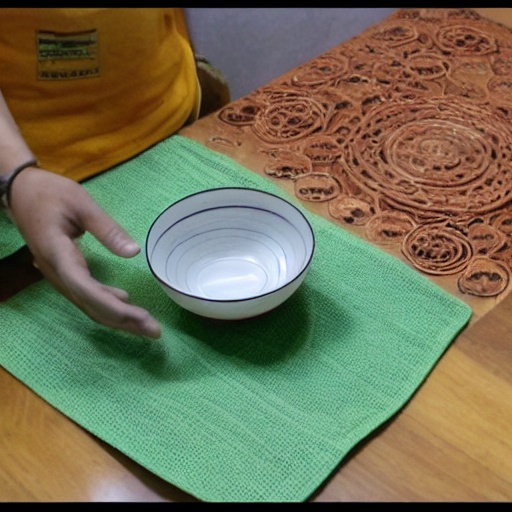} &
         \includegraphics[width=0.125\textwidth]{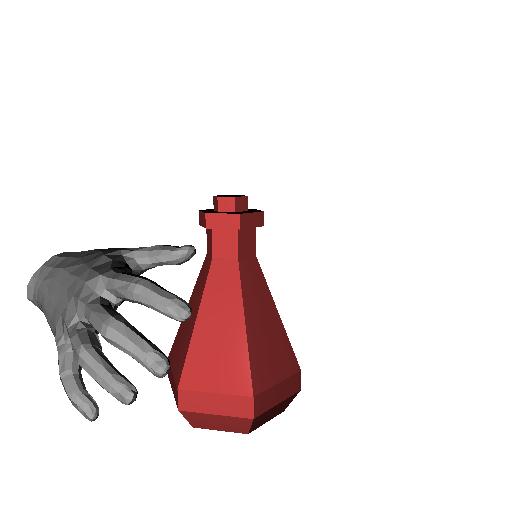}  &
         \includegraphics[width=0.125\textwidth]{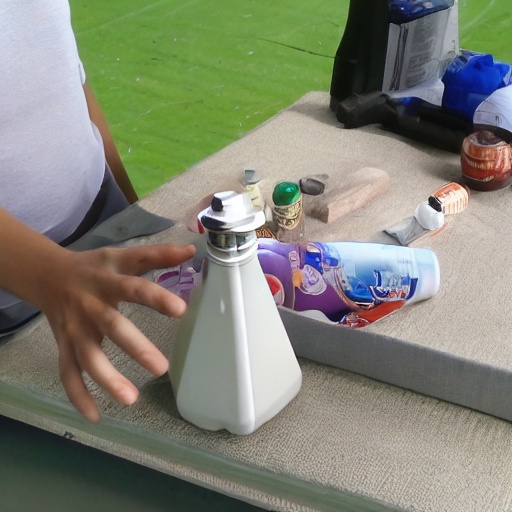} &
         \includegraphics[width=0.125\textwidth]{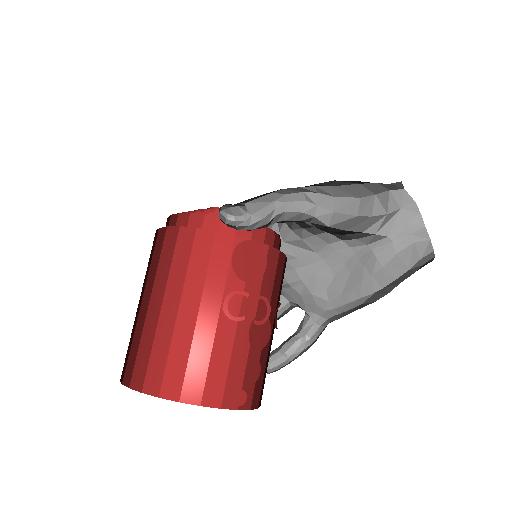}&
         \includegraphics[width=0.125\textwidth]{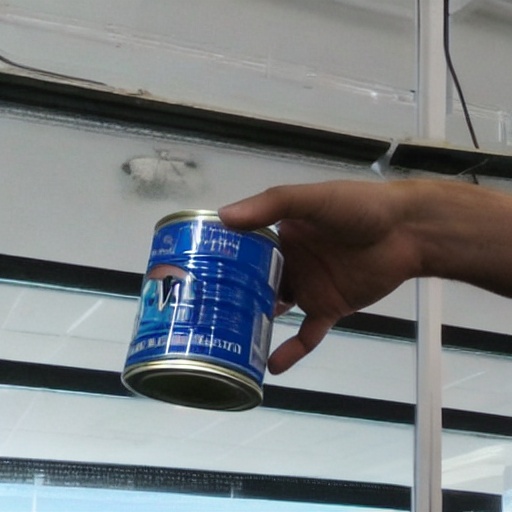} \\
         
        \includegraphics[width=0.125\textwidth]{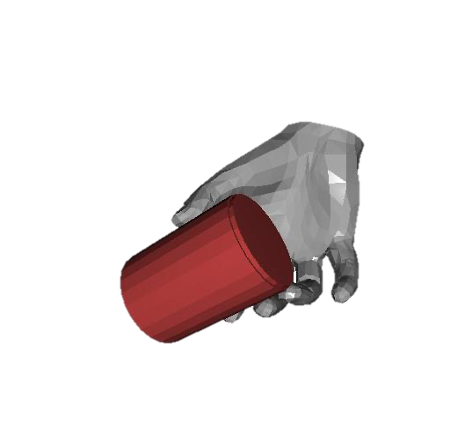} &
         \includegraphics[width=0.125\textwidth]{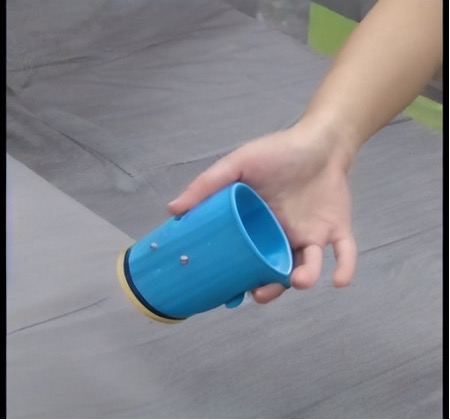}  &
         \includegraphics[width=0.125\textwidth]
         {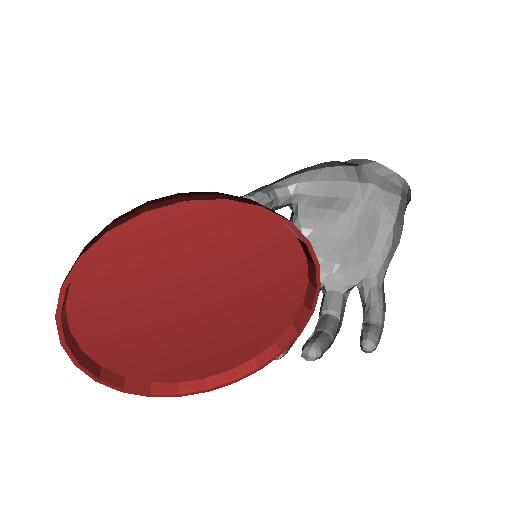}  &
         \includegraphics[width=0.125\textwidth]{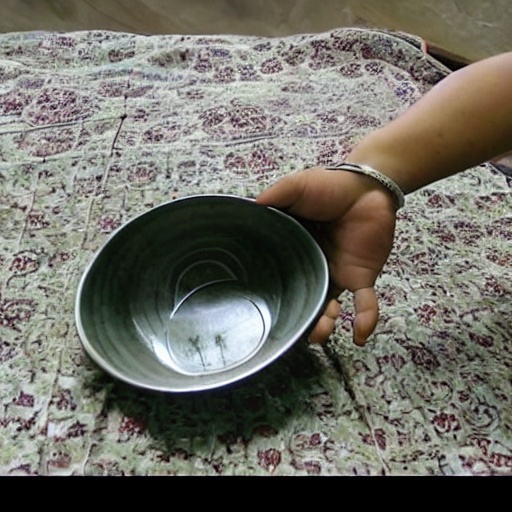} &
         \includegraphics[width=0.125\textwidth]{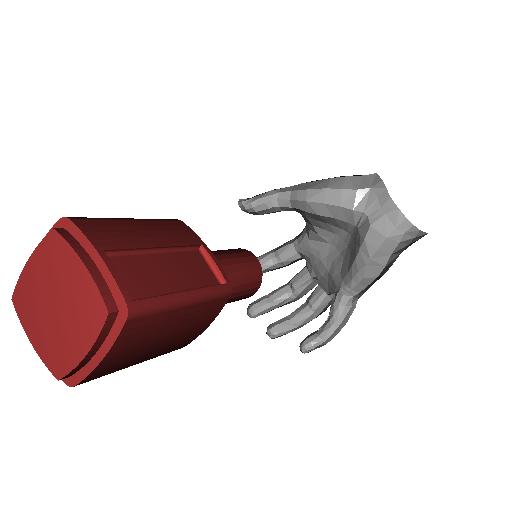}  &
         \includegraphics[width=0.125\textwidth]{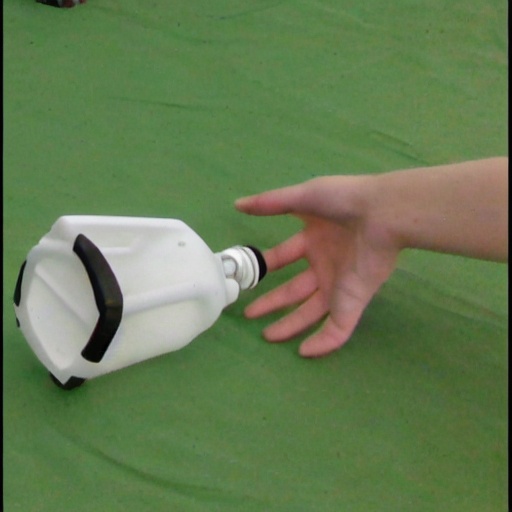} &
         \includegraphics[width=0.125\textwidth]{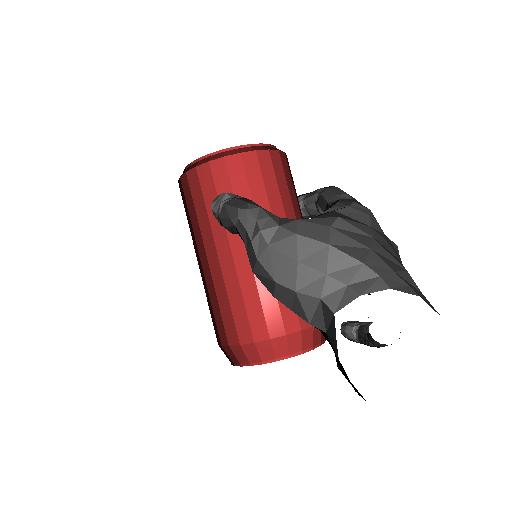}&
         \includegraphics[width=0.125\textwidth]{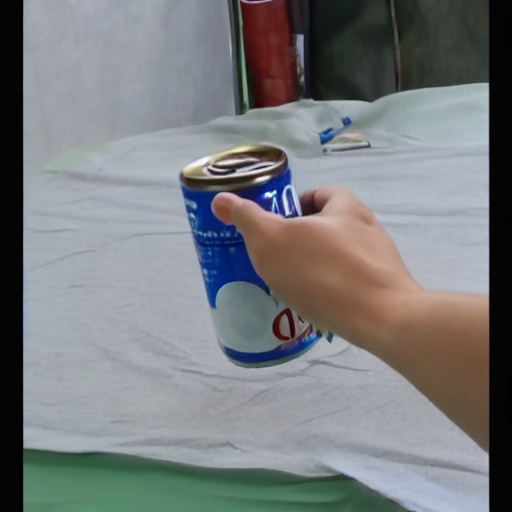} \\
         
         \includegraphics[width=0.125\textwidth]{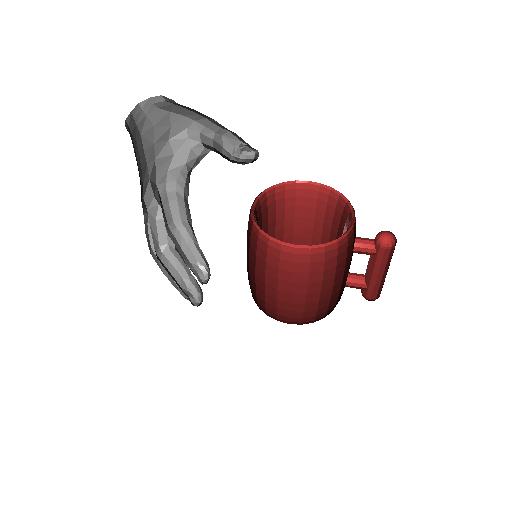} &
         \includegraphics[width=0.125\textwidth]{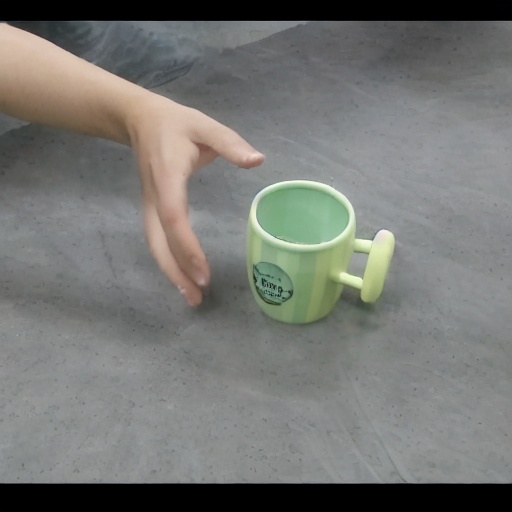}  &
         \includegraphics[width=0.125\textwidth]{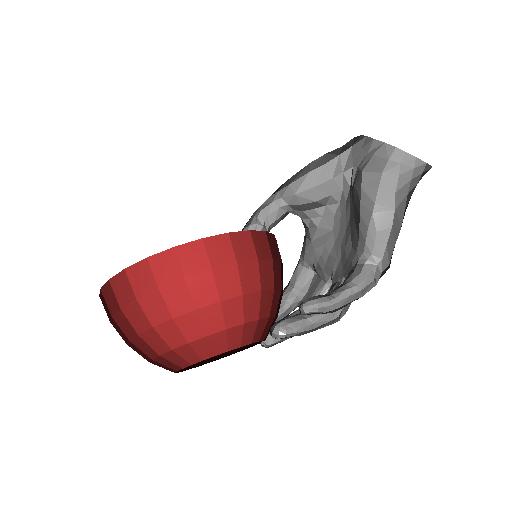}  &
         \includegraphics[width=0.125\textwidth]{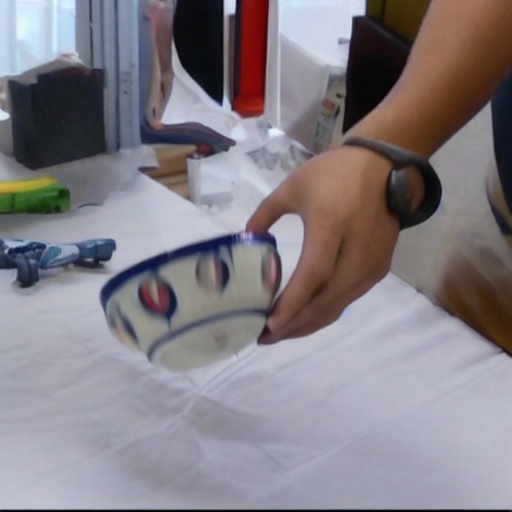} &
         \includegraphics[width=0.125\textwidth]{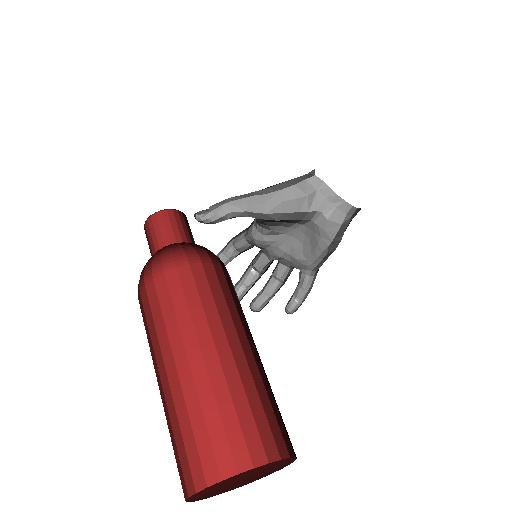}  &
         \includegraphics[width=0.125\textwidth]{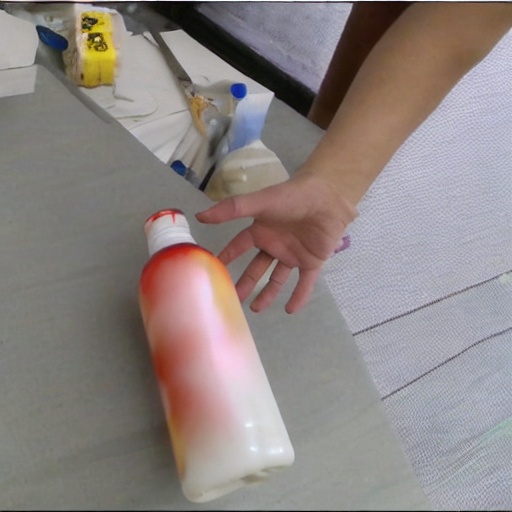} &
         \includegraphics[width=0.125\textwidth]{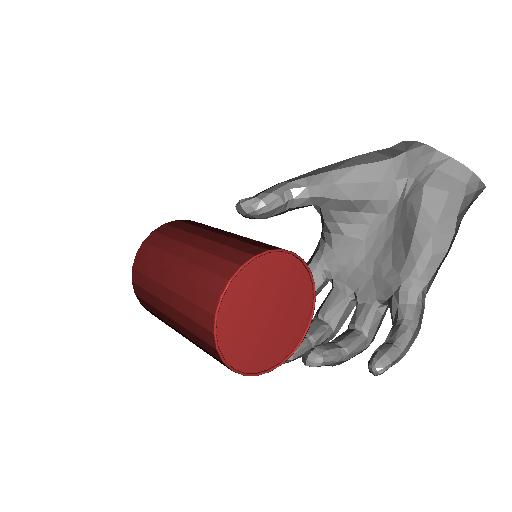}&
         \includegraphics[width=0.125\textwidth]{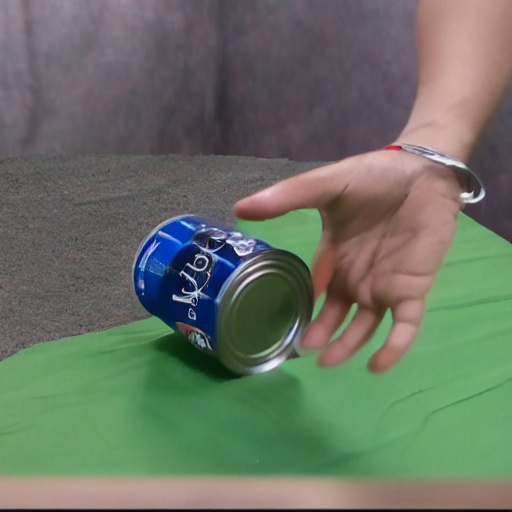}
         \\
    \end{tabular}
    }
    \caption{\textbf{Qualitative results on different structures.} Generated images with the same background description but different physical conditions (object shape, poses, and hand skeletons). With plain prompts, HOIDiffusion could generate more realistic images similar to the style in training datasets.}
    \vspace{-3mm}
    \label{fig:geometry}
\end{figure*}
\noindent \textbf{Hand Pose Evaluation} For hand-object-interaction images, hand grasping status, and pose precision are of vital importance for real-world applications. A fundamental criterion for synthesized images is the geometric consistency between our generated hands and the provided 2D skeleton projection, disregarding the depth dimension. On top of that, these grasping images sometimes serve as visual demonstrations for various downstream tasks, requiring the exactly accurate contact status between the hand and object in the ending pose image along the grasping trajectory. Consequently, our model is evaluated with the baselines on two key perspectives: Hand contact recall and hand re-inference accuracy. Specifically, we adopt a contact evaluation setup utilized in Affordance Diffusion~\cite{ye2023affordance}. An off-the-shelf hand-object detector~\cite{Shan20} is used to classify the image's in-contact status. Furthermore, to evaluate the re-inference accuracy, we estimate the MANO parameters of hands in images through a widely used single-view hand pose estimator~\cite{rong2021frankmocap}, from which we derive the predicted hand joint positions. The percentage of correct keypoints (PCK) is used to measure the accuracy of predicted keypoints representing the hand poses in our data. We evaluate the pose precision on a subset of daily objects. As delineated in Table \ref{tab:hand_eval}, benefited from the geometry guidance, our method manifests the capability to synthesize images depicting accurate grasps and firm contacts, outperforming previous methods in achieving higher mean contact recall and PCK.

\begin{table}[t]
    \vspace{-2mm}
    \centering\small
    
    \scalebox{0.75}{
    \begin{tabular}{l c c c c c c}
         \toprule
          \multirow{2}{3em}{Method}& \multicolumn{5}{c}{Hand Contact Recall \%} & \multirow{2}{3em}{PCK$\uparrow$} \\
          & Mug & Bowl & Bottle & Can & Mean$\uparrow$ & \\
         \midrule
         LDM~\cite{rombach2022high} & 79.12 & 78.89 & 73.00 & 60.64 & 72.91 & 0.15  \\
         DreamBooth~\cite{ruiz2023dreambooth} & 79.12 & 62.22 & 78.00 & 75.53 & 73.72 & 0.10\\
         ControlNet~\cite{zhang2023adding} & 86.00 & 91.40 & 89.00 & 93.00 & 89.85 & 0.67\\
         Affordance Diffusion~\cite{ye2023affordance} & 73.00 & 90.00 & 90.00 & - & 84.33 & - \\
         \midrule
          HOIDiffusion (Ours) & 92.31 & 97.78 & 94.00 & 97.87 & \textbf{95.49} & \textbf{0.85} \\
    
         \bottomrule
    \end{tabular}
    }
    \vspace{-3mm}
    \caption{\small {\textbf{Evaluation metrics from hand perspective.} Hand Contact Recall is to evaluate whether the hand-ending pose is in close contact with the object. PCK is used to measure the accuracy of generated images' keypoints 
    }}
    \label{tab:hand_eval}
    \vspace{-5mm}
\end{table}

\subsection{Geometry and Appearance Disentangle}
The most important design in our model is to disentangle the physical geometries from textures. Through this design, we have observed the remarkable ability of our model to control the generation process with novel object shapes and previously unseen text descriptions. During training, our model learns extensive shape priors from masks and normal map conditions, hence acquiring the capability to transfer to unseen object instances seamlessly. Furthermore, our model preserves the robust text editing ability, enabling flexible style transformation over both background and object appearance. In this subsection, we primarily focus on and showcase the qualitative results of structure and style manipulation.

\noindent \textbf{Geometry Manipulation}
In Figure \ref{fig:geometry}, we present the generated paired images of four daily seen objects: mug, can, bottle, and bowl, given varying instance shapes, poses, and hand skeletons. The left column of each pair is rendered image in 3D space, from which physical conditions are extracted. Normal maps are obtained from an estimated depth provided by the depth estimator MiDaS~\cite{Ranftl2022}. The skeleton and segmentation are also concurrently obtained during rendering. Corresponding generated images are displayed in the right column. All provided prompts follow the format: "A hand is grasping a [object]". The results exhibit an overall generation style in a laboratory or stereo environment, consistent with the realistic appearance style in the training dataset. From Figure \ref{fig:teaser}, it is also evident that HOI images generated from our model are more closely aligned with the required geometry than baseline methods.

\begin{figure*}
\centering
\scalebox{0.95}{
    \setlength\tabcolsep{0.5pt}
    \renewcommand{\arraystretch}{0.25}
    \begin{tabular}{cccccc}
         \includegraphics[width=0.167\textwidth]{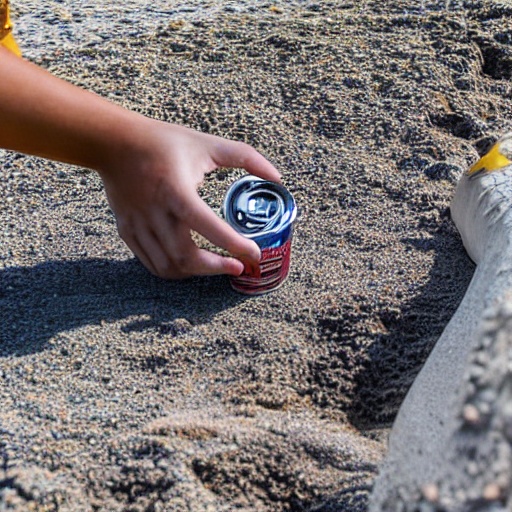} &
         \includegraphics[width=0.167\textwidth]{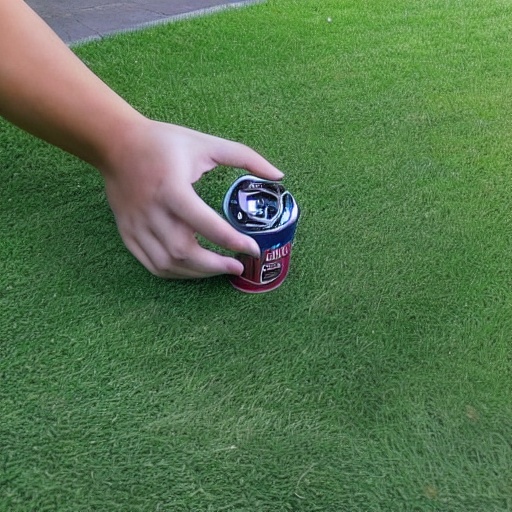}  &
         \includegraphics[width=0.167\textwidth]{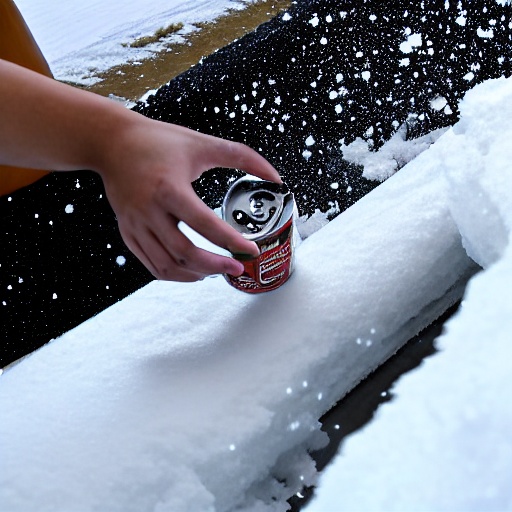}  &
         \includegraphics[width=0.167\textwidth]{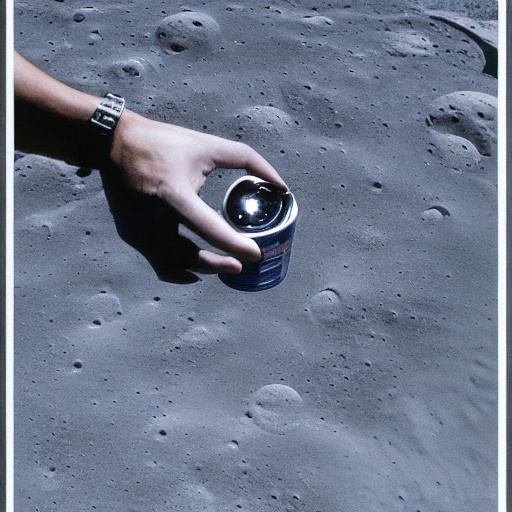} &
         \includegraphics[width=0.167\textwidth]{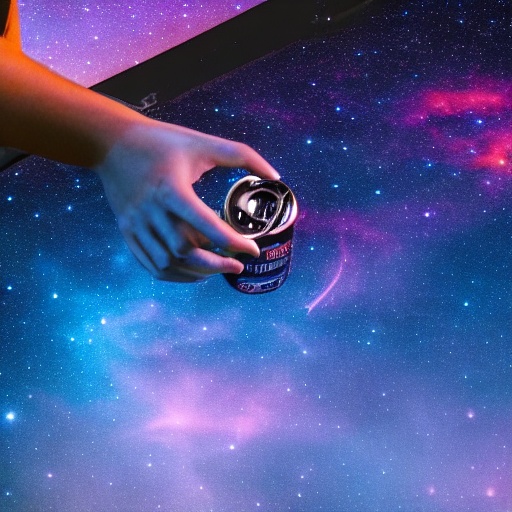}  &
         \includegraphics[width=0.167\textwidth]{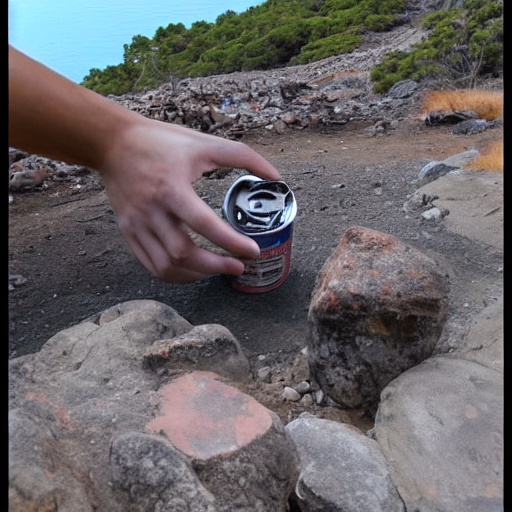} \\
         
        \includegraphics[width=0.167\textwidth]{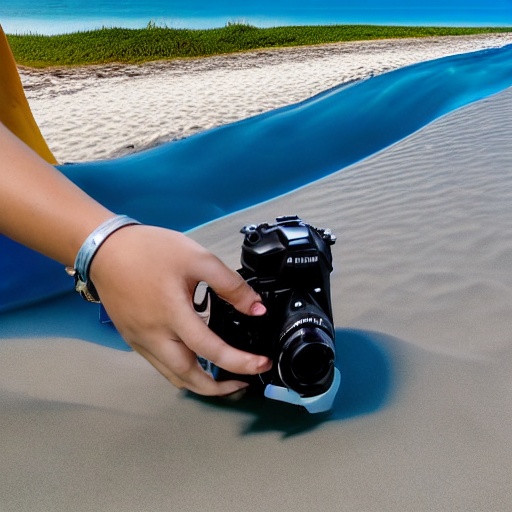} &
         \includegraphics[width=0.167\textwidth]{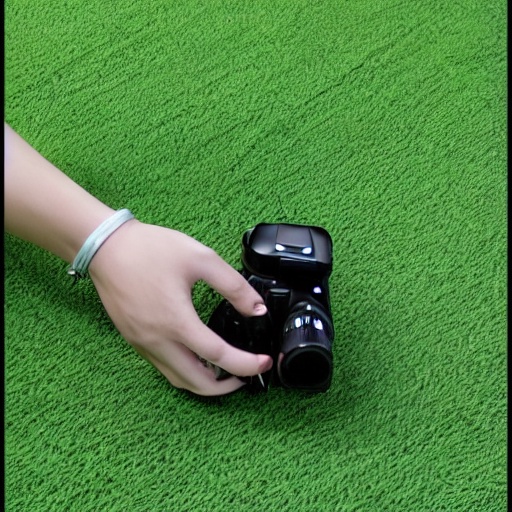}  &
         \includegraphics[width=0.167\textwidth]{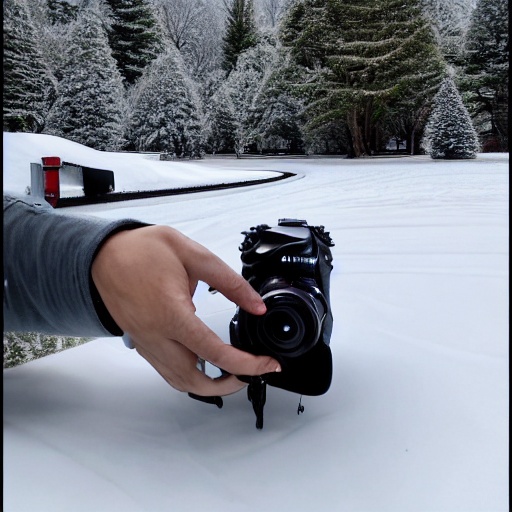}  &
         \includegraphics[width=0.167\textwidth]{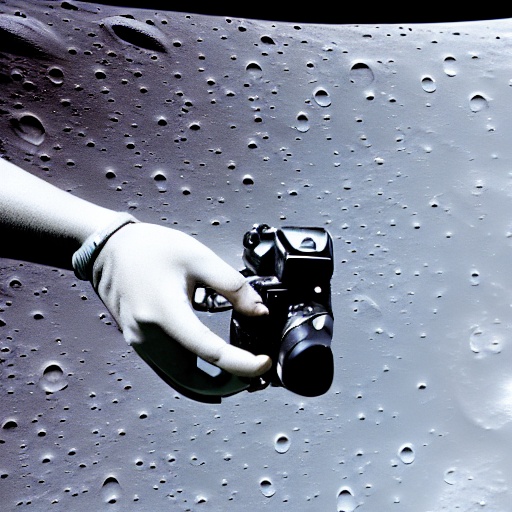} &
         \includegraphics[width=0.167\textwidth]{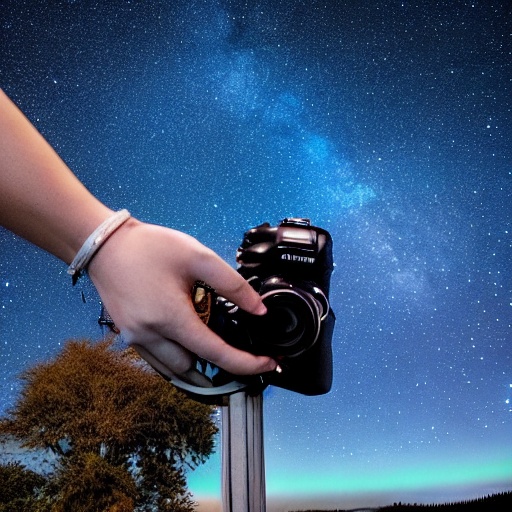}  &
         \includegraphics[width=0.167\textwidth]{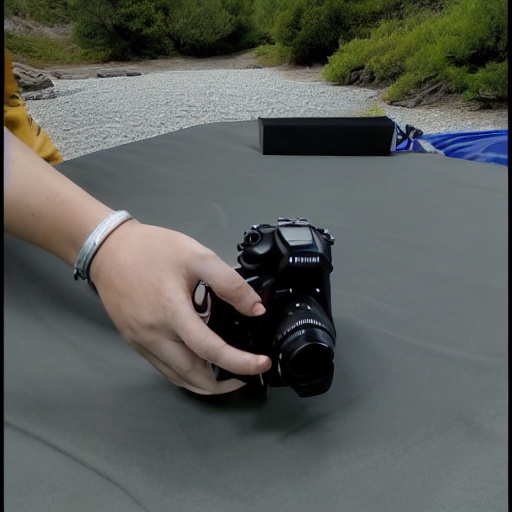}  \\
         
         \includegraphics[width=0.167\textwidth]{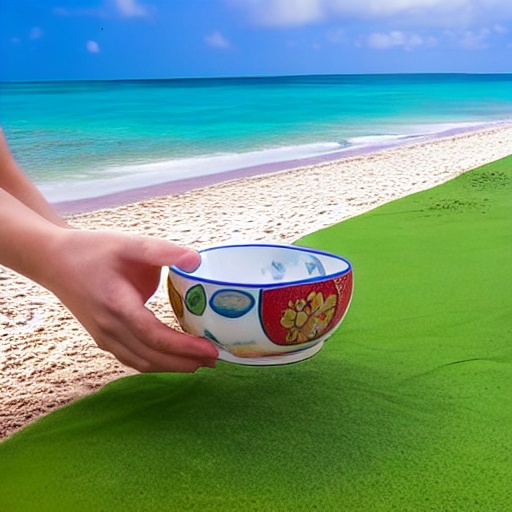} &
         \includegraphics[width=0.167\textwidth]{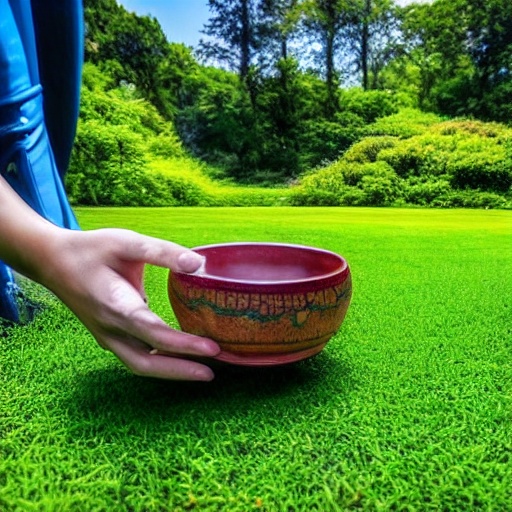}  &
         \includegraphics[width=0.167\textwidth]{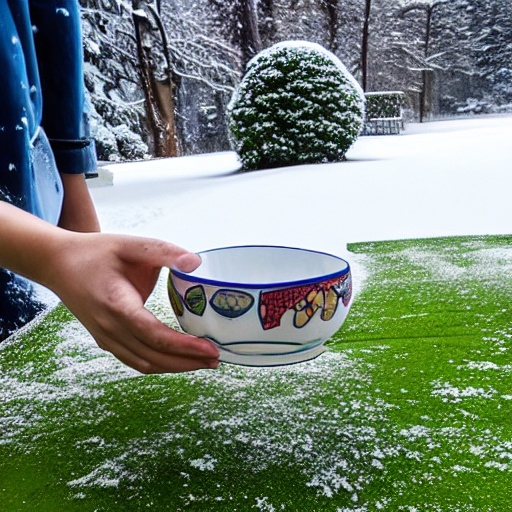}  &
         \includegraphics[width=0.167\textwidth]{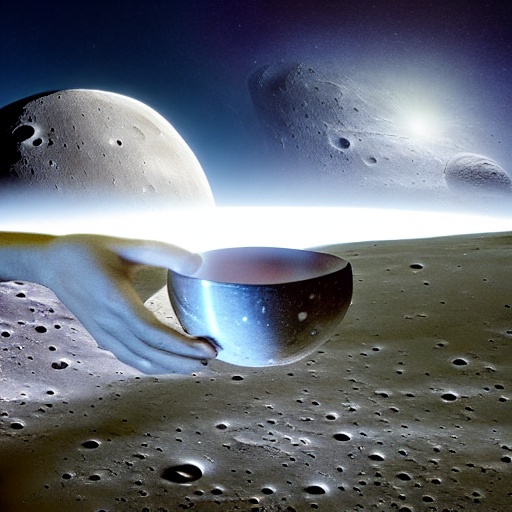} &
         \includegraphics[width=0.167\textwidth]{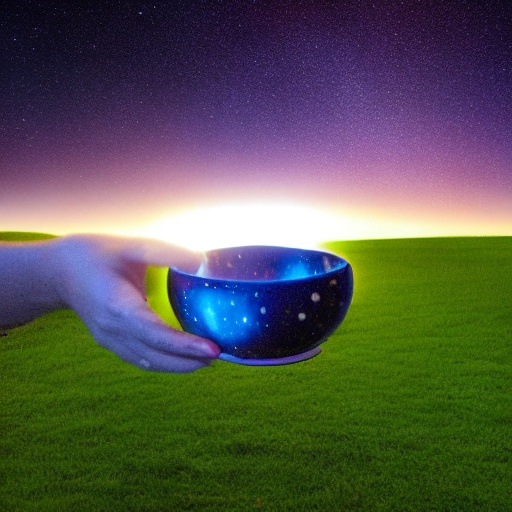}  &
         \includegraphics[width=0.167\textwidth]{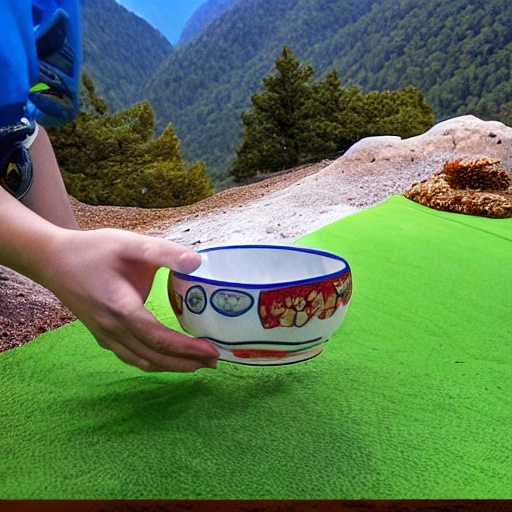}
         \\
         \\
         on the beach & on the grass & in the snow & on the moon & with night sky & in the mountain
    \end{tabular}
    }
    \vspace{-3mm}
    \caption{\textbf{Synthesized images with diverse background descriptions.} In addition to real-style synthesis, our model also allows users to generate according to their preferences such as science fiction or general landscapes.}
    \vspace{-4mm}
    \label{fig:background}
\end{figure*}
\noindent\textbf{Background and object appearance control} In this section, we explore the text editing ability with fixed geometry. Through background regularization and classifier-free guidance, our model exhibits the ability to depict diverse background contents, retaining control over appearance using text prompts. 
We investigate the qualitative performance of various text controls under identical physical conditions, shown in Figure \ref{fig:background}. Each column represents one distinct background description. Notably, the generated images exhibit high fidelity to the provided prompts, maintaining the layout and structure unchanged. This demonstrates the ability of HOIDiffusion to effectively disentangle the appearance from geometry structures, thus enabling flexible style transformation without geometry distortion. This is essential in data construction, ensuring the precise alignment with input geometry and diverse range of visual appearances.

\subsection{Applications}
\begin{figure*}
\centering
\scalebox{0.95}{
    \setlength\tabcolsep{0.5pt}
    \renewcommand{\arraystretch}{0.25}
    \begin{tabular}{c}
         \includegraphics[width=1\textwidth]{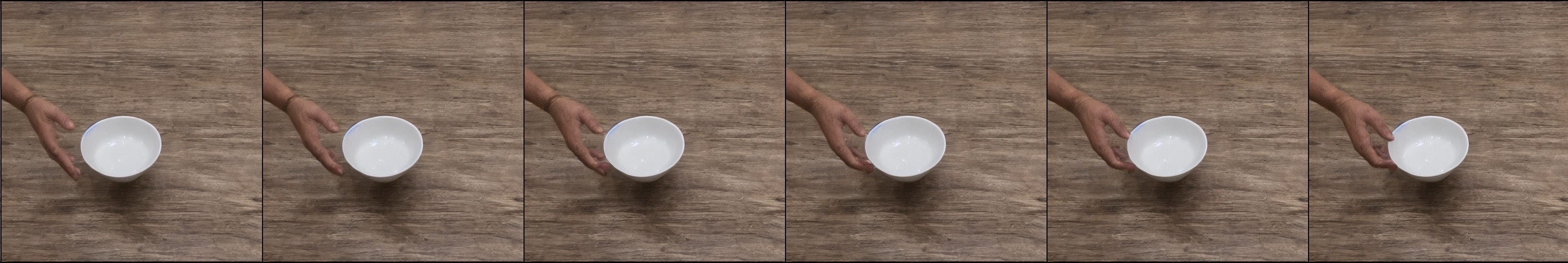}
          \\
        \includegraphics[width=1\textwidth]{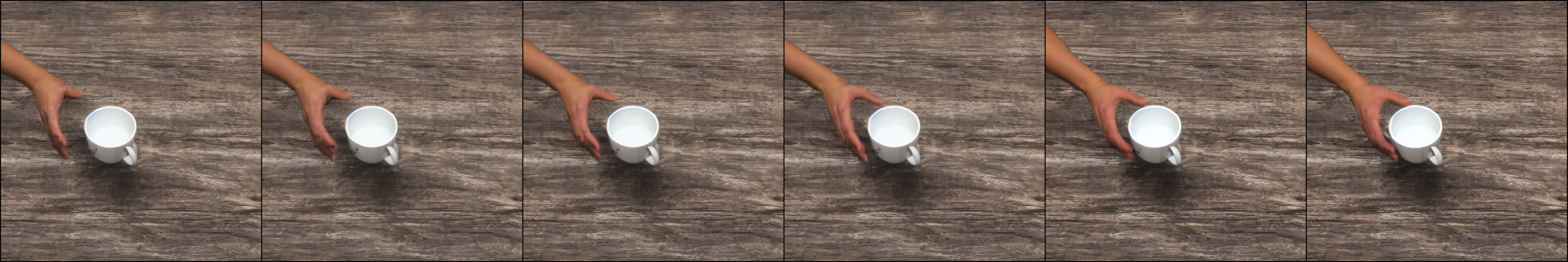} 
    \end{tabular}
    }
    \vspace{-2mm}
    \caption{\textbf{Zero-shot video generation of hand grasping trajectory.} Images along the same line represent the sequential motion of reaching an object. By leveraging temporal-level cross-attention, the frame flickering problem is mitigated.}
    \label{fig:video}
\end{figure*}

\noindent\textbf{Video Generation}  The real hand-object interaction datasets often exhibit data in video format, a complete fetching process to the object. Collecting these video clips is a considerable challenge. Some datasets~\cite{chao2021dexycb, yang2022oakink} comprising almost millions of images only contain no more than 10k videos, thus video data is much more valuable. In experiments, we observe the significant divergence in generated images between adjacent frames despite the similar provided conditions and texts. This divergence results in dramatic flickering in directly concatenated videos. To this end, we leverage zero-shot video generation techniques in the diffusion model to establish inter-consistency among frames. To be more specific, the original self-attention layers in the U-Net are refactored to cross-attention modules between an anchor frame and current frames, This adjustment establishes the awareness of the previous appearance style, ensuring video consistency. In our experiments, we set the middle frame as an anchor, and with all frames attend to both the anchor image and themselves. This approach effectively mitigates the flickering issue, synthesizing relatively smooth hand-grasping trajectories. The video clip samples are shown in Figure \ref{fig:video}.

\begin{table*}[t]
    \centering\small
    
    \scalebox{0.87}{
    \begin{tabular}{l l c c c c c c c}
         \toprule
          \multicolumn{2}{c}{Method}& IoU@25 & IoU@50 &IoU@75 &5$^\circ$2cm & 5$^\circ$5cm &10$^\circ$2cm&10$^\circ$5cm\\
         \midrule
         \multirow{2}{5em}{SPD\\\cite{a-sdf}}&Original & \textbf{82.9} & \textbf{75.3} & \textbf{48.6} & 17.7 & 19.9 & 38.8 & 48.3  \\
         &Ours& 82.5& 71.1 &47.0 &\textbf{21.1}&\textbf{23.2}&\textbf{43.8} &\textbf{54.5}\\
         \midrule
         \multirow{2}{5em}{DualPoseNet ~\cite{lin2021dualposenet}} &Original&84.1 &79.7 & 60.1 & 28.0 & 34.3 & 47.8 & 64.2\\
         &Ours& \textbf{90.9}& \textbf{84.1} &\textbf{65.8} &\textbf{29.2}&\textbf{34.4}&\textbf{55.0} &\textbf{66.6}\\
         \bottomrule
    \end{tabular}
    }
    \vspace{-2mm}
    \caption{\small {\textbf{Quantitative evaluation on NOCS.} We use SPD and DualPoseNet and change the synthesized images in the dataset with our generated images for training. Our performance improve on all metrics with DualPoseNet and all cm metrics with SPD which demonstrates the good quality of our images and can be utilized for downstream tasks. }}
    \label{tab:obj_eval}
    \vspace{-2mm}
\end{table*}

\noindent\textbf{Downstream tasks} Another interesting method to evaluate the performance of HOIDiffusion, is to apply it in downstream tasks as a data augmentation method or new data source. In this section, we explore the potential for improving categorical object 6D pose estimation tasks. Most models in this task are trained on dataset NOCS ~\cite{wang2019normalized}, consisting of both synthetic and real data with annotations.  Some models~\cite{Tian_2020_ECCV,chen2021sgpa} implicitly predict normalized object coordinate space, and then utilize Umeyama algorithms to parse the transformation matrix. Others~\cite{lin2021dualposenet} explicitly predict the rotation, translation, and scale parameters. Despite differing in module design, all these methods leverage RGB image encoders to obtain the visual features. An interesting observation is that the synthesized images directly rendered from object models appear too artificial, potentially affecting the performance of the RGB encoder. Inspired by this, we substitute the synthesized images with our generated HOI images in the same poses, anticipating the reality brought by our data could help enhance model performance. We exhibit the results in Table \ref{tab:obj_eval}. We choose two representative object pose estimators for evaluation. The “original” model in the table refers to training using an unchanged NOCS dataset, and “our” model is trained using our mixed data.

\vspace{-0.8em}
\section{Ablation Study}
Two indispensable components of our design are   
precise structural control and appearance regularization, effectively improving model performance on geometry consistency and diversity.

\noindent \textbf{Structural Control}  In our approach, there are three crucial structure conditions provided to the model and we investigate the importance brought separately by these three modules. Results are presented in Table \ref{tab:ablation_cond}. Essentially, each condition serves a distinct purpose: the normal map guides the model in perceiving surface textures with lighting, which is essential for maintaining geometry consistency; hand keypoint projection precisely depicts the pose of hand joints, preventing the model from synthesizing multiple fingers or distorted hands; hand-object segmentation provides a clear boundary between different regions, avoiding interference between hand and object areas. As presented in the results, our full model outperforms other incomplete versions in quantitative evaluation.   

\begin{table}[!ht]
    \centering
    \scalebox{0.8}{
    \begin{tabular}{l c}
    \toprule
    Method & FID (1k) $\downarrow$ \\
    \midrule
    LDM (finetuned) & 86.12 \\
    w/o estimated normal maps & 82.40 \\
    w/o hand keypoint projection & 81.93 \\
    w/o hand-object segmentation & 78.57 \\
         \midrule
    HOIDiffusion(Ours) & \textbf{77.64} \\
    \bottomrule
    \end{tabular}}
    \vspace{-2mm}
    \caption{\small \textbf{Ablation study on structural control.} FID evaluation on 1,000 images of different types of missing modules to demonstrate the necessity of all physical conditions. Our method outperforms all others.}
    \vspace{-2mm}
    \label{tab:ablation_cond}
\end{table}

\begin{table}[!ht]
    \centering
    \vspace{-3mm}
    \scalebox{0.8}{
    \begin{tabular}{l c}
    \toprule
    Method & CLIPScore $\uparrow$ \\
    \midrule
    w/o regularization & 0.66 \\
    HOIDiffusion(Ours) & \textbf{0.79} \\
    \bottomrule
    \end{tabular}}
    \vspace{-2mm}
    \caption{\small \textbf{CLIPScore evaluation.} Consistency is evaluated between provided prompts and generated images for different backgrounds and instances.}
    \vspace{-4mm}
    \label{tab:ablation_reg}
\end{table}

\begin{figure}[!ht]
\centering
\scalebox{0.95}{
    \begin{tabular}{c}
\includegraphics[width=0.48\textwidth]{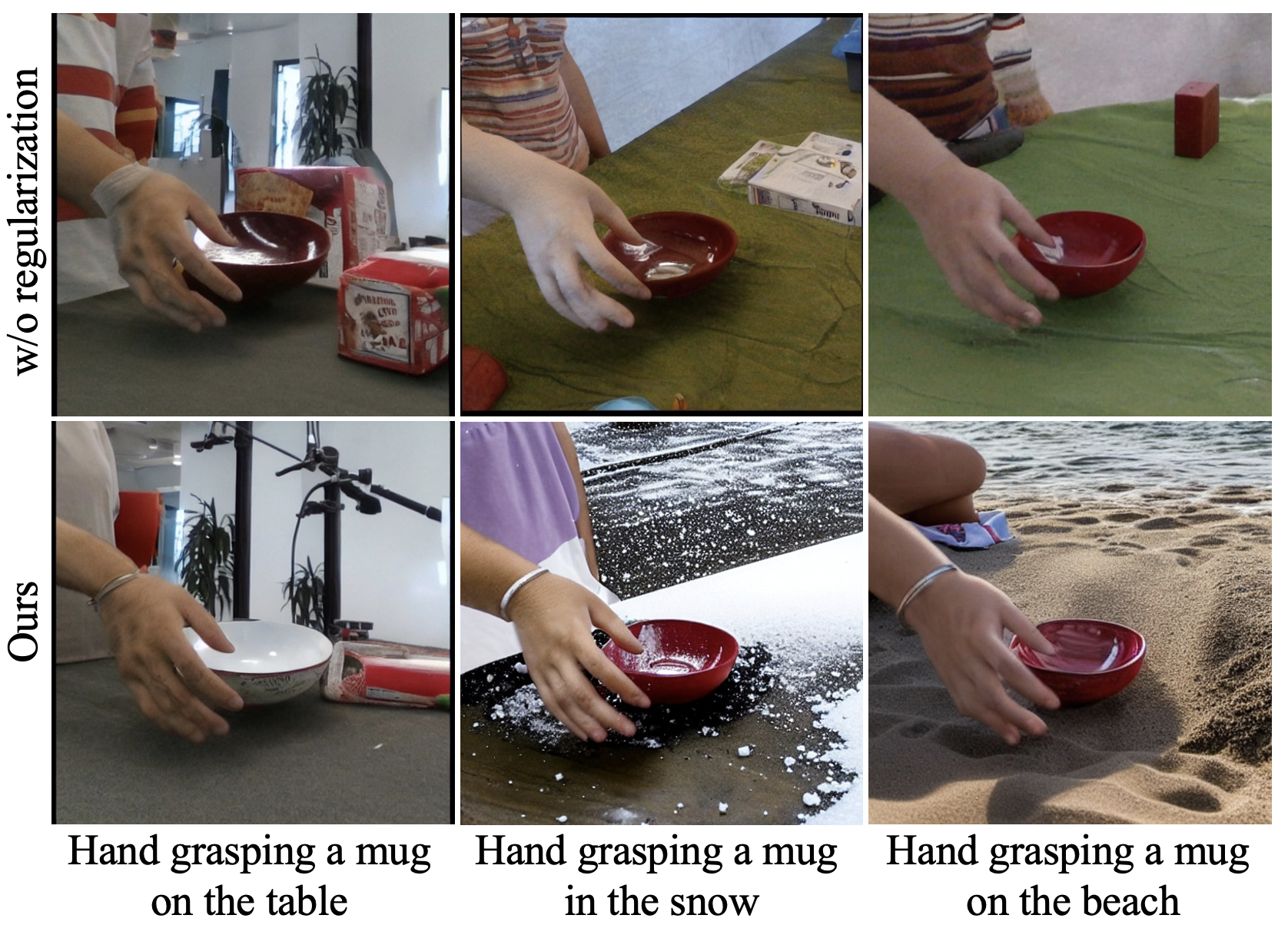}\\

\end{tabular}
}
\vspace{-2mm}
\caption{\small \textbf{Ablation study on appearance regularization}. Different backgrounds are used as prompts with the same geometry conditions to compare the text editing flexibility brought by the regularization module.}
    \label{fig:ablation_reg}
\end{figure}

\noindent \textbf{Appearance Regularization}  Table \ref{tab:ablation_reg} and Figure \ref{fig:ablation_reg} demonstrate the significance of appearance regularization in our module. Without regularization, the finetuned model quickly converges to the style in training datasets, mostly in a laboratory/studio environment, as depicted in the first line of Figure \ref{fig:ablation_reg}, This convergence impairs the model's ability to generate diverse images, which is essential for data generation. By incorporating appearance regularization using data generated from the pretrained model, HOIDiffusion mitigates the drift to a fixed style and improves the overall text-editing ability.

\vspace{-4mm}
\vspace{-4mm}

\vspace{-0.1in}
\section{Conclusion}
\vspace{-0.05in}
In this paper, we propose HOIDiffusion with precise appearance and structure control. We did experiments on geometry and appearance manipulation, and evaluated the performance using FID, IS, and hand contact recall. The results demonstrate better performance compared to baseline models. We apply the generated data for object 6D pose estimation and show its effectiveness in possibilities to improve perception systems.
{
    \small
    \bibliographystyle{ieeenat_fullname}
    \bibliography{main}
}
\clearpage
\pagenumbering{arabic}
\renewcommand*{\thepage}{A\arabic{page}}

\appendix

\newcommand{\appendixhead}
{\centering{\Large \bf Appendix}
\vspace{10mm}}

\twocolumn[\appendixhead]
\section{Architecture Details}
\label{sec:sup_arc}
  
There are two components in HOIDiffusion: the main diffusion model and the condition model. For the main branch, we base our diffusion model on Stable Diffusion~\cite{rombach2022high} with changes of taking condition embeddings from three physical sources: normal map, hand skeleton projection, and segmentation. The conditions are added into the U-Net encoder, to shift feature maps of each encoding channel, similar to the method adopted in Adapter~\cite{mou2023t2i}. During training, we turn on the training of pretrained Stable Diffusion's decoder, with other parameters fixed (including other modules in U-Net and the text encoder). Model parameter details are provided in Table \ref{tab:main_param}. 

For condition models, we encode the condition images into different embedding levels for all feature channels in the diffusion model encoder. We adopt ResNet blocks to obtain the embeddings for each channel. Since there are three structural physical sources, we adopt the same model architecture for different conditions and utilize a weighted sum as the final condition outputs. Detailed information is shown in Table \ref{tab:cond_param}.  
\begin{table}[!htbp]
    \centering
    \scalebox{0.85}{
    \begin{tabular}{l c}
    \toprule
    Parameter &  Diffusion Model (512$\times$512)\\
    \midrule
    Latent Shape & 4$\times$64$\times$64 \\
    Channels & 320\\
    Channels Multiple & [1, 2, 4, 4] \\
    ResBlock Number & 2 \\
    Context Dimension & 768 \\ 
    Batch Size & 8 \\
    Diffusion Steps & 1000\\
    Noise Scheduler & Linear\\
    Learning Rate & 10$^{-5}$ \\
    Optimizer & Adam\\
    \bottomrule
    \end{tabular}}
    \caption{Model architecture and training scheme (decoder) for main diffusion models. Input images are all resized to 512$\times$512.}
    \label{tab:main_param}
    \vspace{-3mm}
\end{table}

\begin{table}[!htbp]
    \centering
    \scalebox{0.85}{
    \begin{tabular}{l c}
    \toprule
    Parameter &  Condition Model (512$\times$512)\\
    \midrule
    Input Channels & 3$\times$64 \\
    Output Channels & [320, 640, 1280, 1280]\\
    ResBlock Number & 2 \\
    Feature Weight (h,n,s) & [1,1,1] \\
    Kernel Size & 1 \\ 
    Batch Size & 8 \\
    Learning Rate & 10$^{-5}$ \\
    Optimizer & Adam\\
    \bottomrule
    \end{tabular}}
    \caption{Model architecture and training hyperparameters for condition model. (h,n,s) in Feature Weight represents summation weights for three conditions: (hand projection, normal map, segmentation).}
    \label{tab:cond_param}
    \vspace{-2mm}
\end{table}
\section{More Results on HOI Generation}
\label{sec:add_res}
We provide more results on realistic hand-object-interaction image generation, with more diverse hand poses, object shapes, and object categories in Figure \ref{fig:sup_hoi}.

\section{Object Appearance Control}
\label{sec:appear}
\begin{figure}[!htbp]
\centering
\scalebox{1}{
    \setlength\tabcolsep{0.5pt}
    \begin{tabular}{cccc}
        \multicolumn{4}{c}{A hand is grasping a pink [object]}\\
         \includegraphics[width=0.125\textwidth]{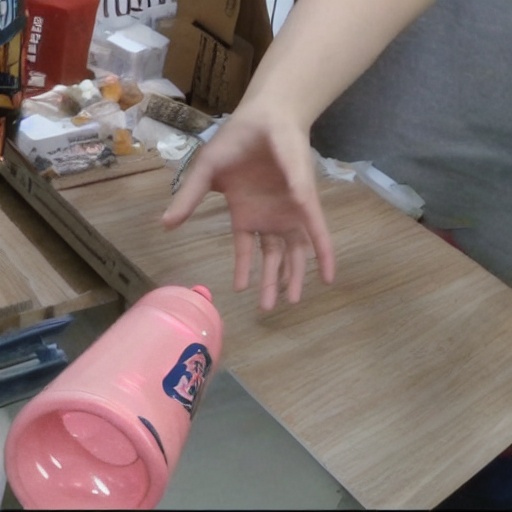} &
         \includegraphics[width=0.125\textwidth]{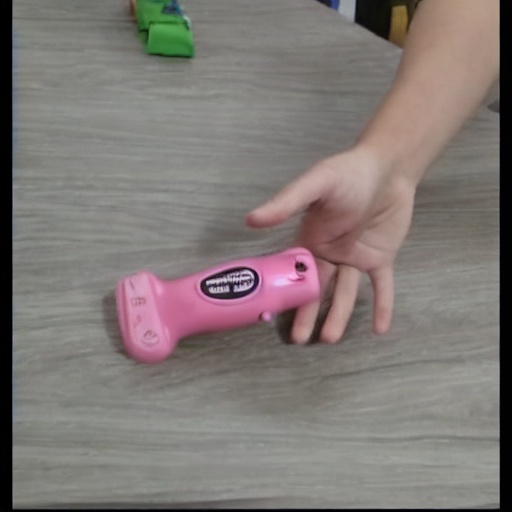}  &
         \includegraphics[width=0.125\textwidth]{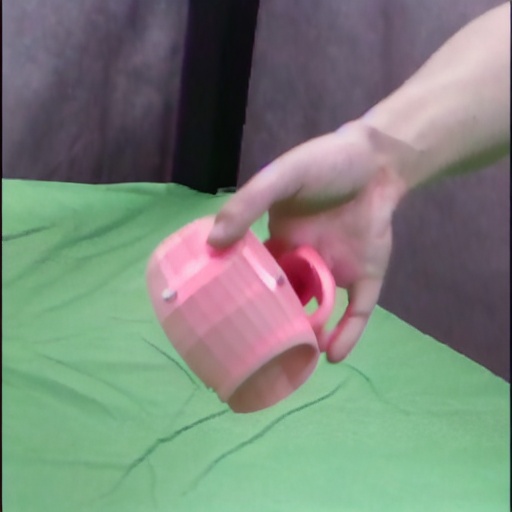}  &
         \includegraphics[width=0.125\textwidth]{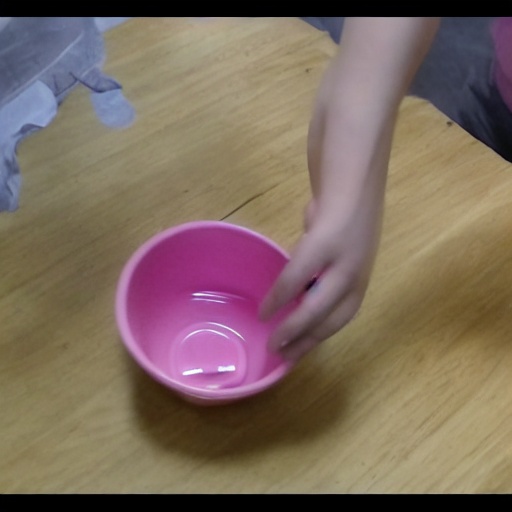} 
         \\
         \multicolumn{4}{c}{A hand is grasping a metallic [object]}\\
         \includegraphics[width=0.125\textwidth]{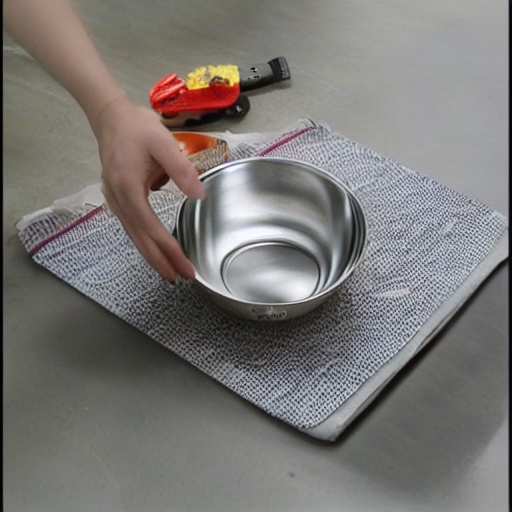} &
         \includegraphics[width=0.125\textwidth]{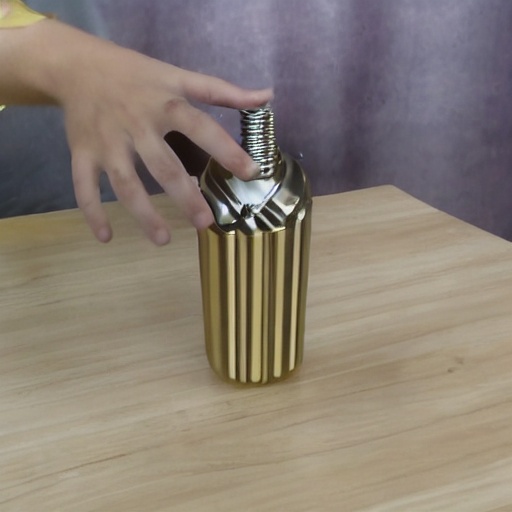}  &
         \includegraphics[width=0.125\textwidth]{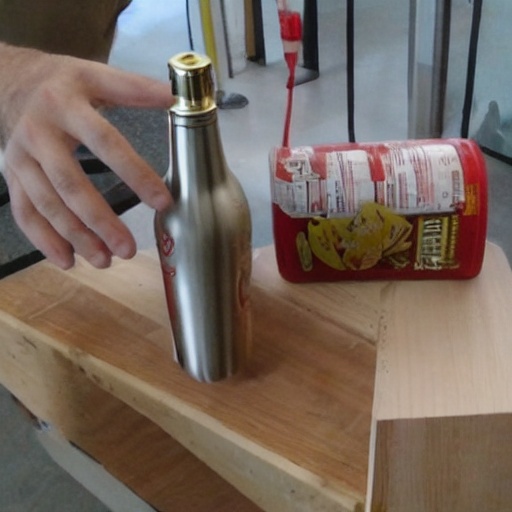}  &
         \includegraphics[width=0.125\textwidth]{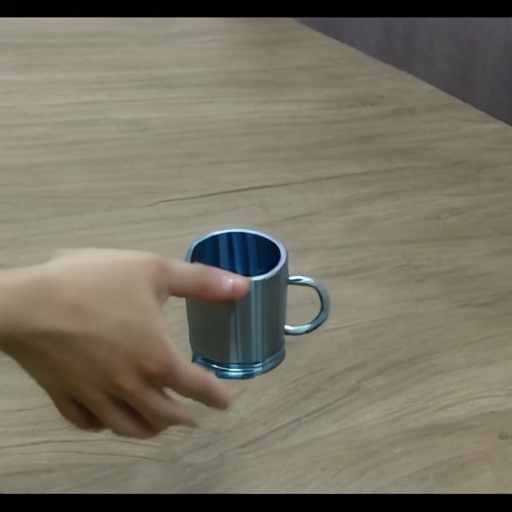} \\
         \multicolumn{4}{c}{A hand is grasping a Gothic-style [object]}\\
         \includegraphics[width=0.125\textwidth]{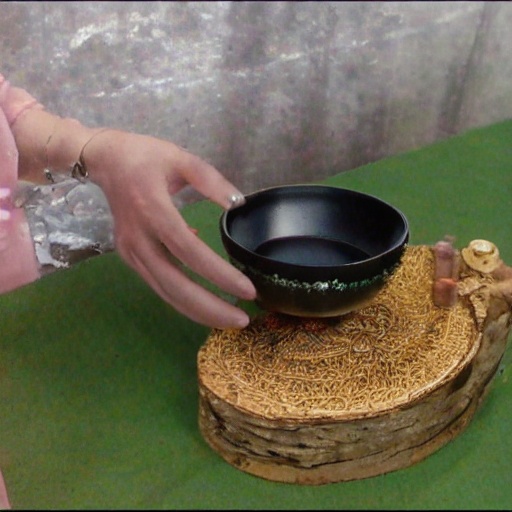} &
         \includegraphics[width=0.125\textwidth]{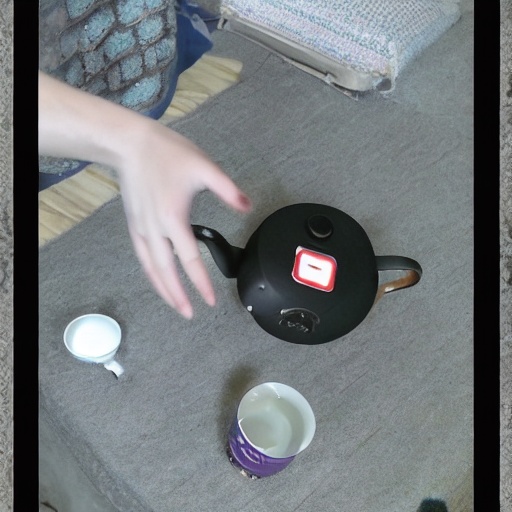}  &
         \includegraphics[width=0.125\textwidth]{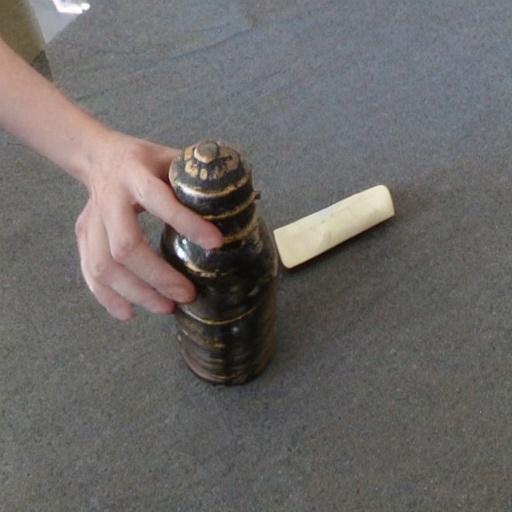}  &
         \includegraphics[width=0.125\textwidth]{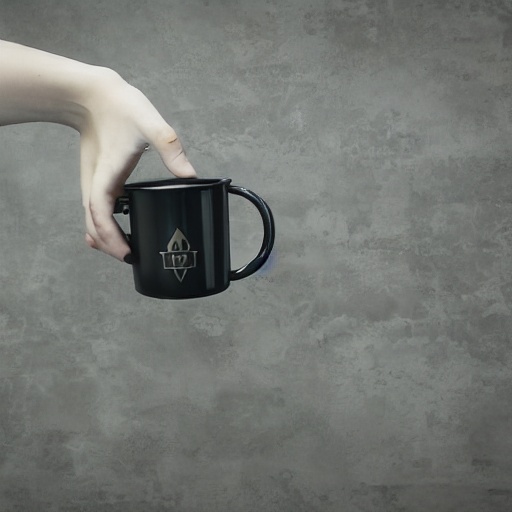} \\
          \multicolumn{4}{c}{A hand is grasping a Medieval-style [object]}\\
         \includegraphics[width=0.125\textwidth]{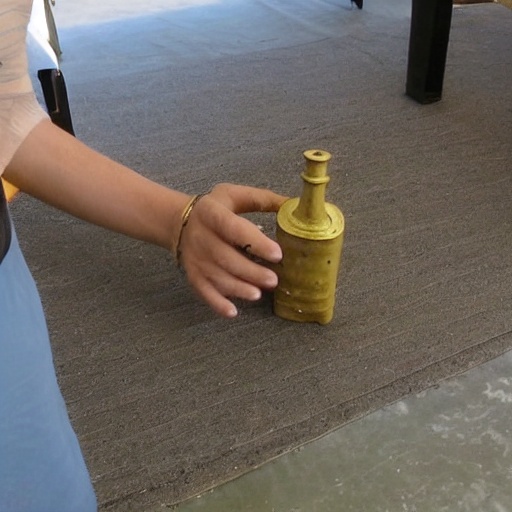} &
         \includegraphics[width=0.125\textwidth]{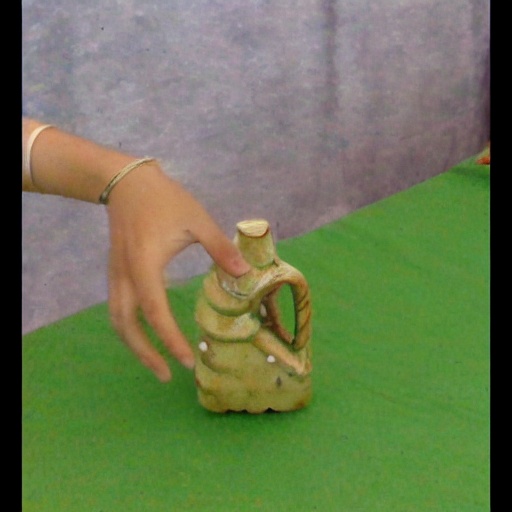}  &
         \includegraphics[width=0.125\textwidth]{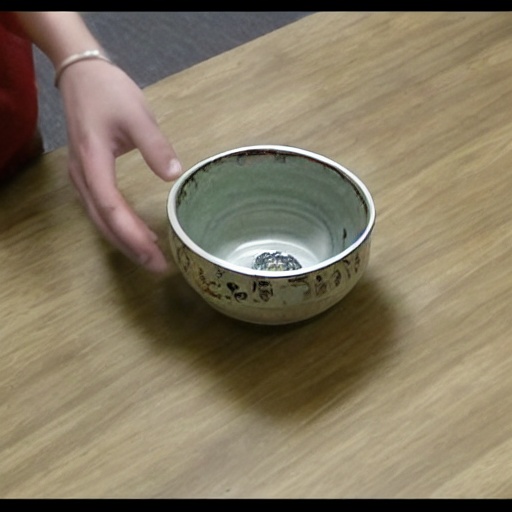}  &
         \includegraphics[width=0.125\textwidth]{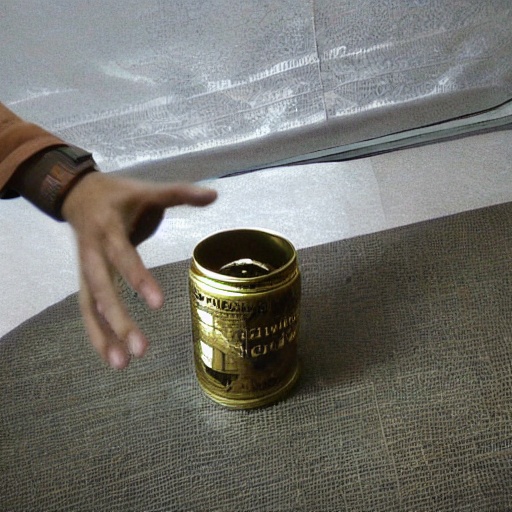} \\
    \end{tabular}
    }
    \vspace{-3mm}
    \caption{Generated images using different style texts to control object appearance.}
    \vspace{-4mm}
    \label{fig:sup_color}
\end{figure}

In this section, we further explore the ability of HOIDiffusion to control the appearance of objects, with different colors or styles unseen in training data or descriptions. The results are shown in Figure \ref{fig:sup_color}. The results can demonstrate that our proposed HOIDiffusion is able to utilize the knowledge from the pretrained models, hence, when a novel appearance is provided, our model is still able to generate expected objects.

\begin{figure*}[t]
\centering
\scalebox{1}{
    \setlength\tabcolsep{0.5pt}
    \renewcommand{\arraystretch}{0.25}
    \begin{tabular}{cccccccc}
    \multicolumn{4}{c}{On a tropical island}&\multicolumn{4}{c}{In the sand}\\
         \includegraphics[width=0.125\textwidth]{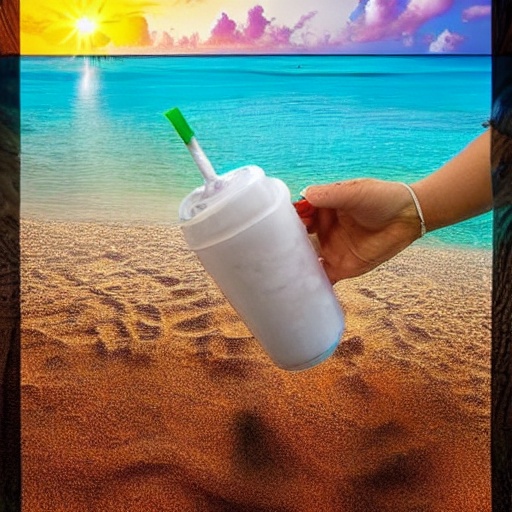} &
         \includegraphics[width=0.125\textwidth]{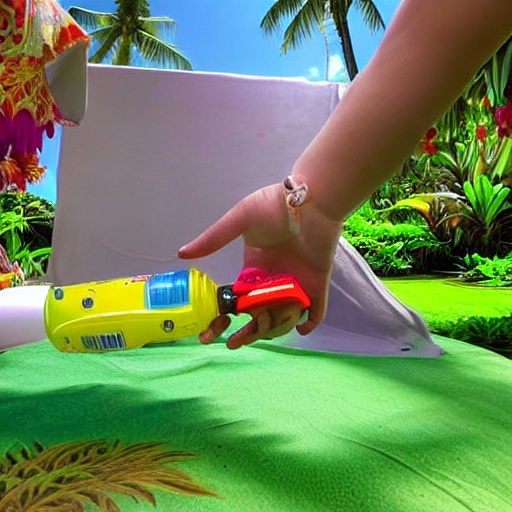}  &
         \includegraphics[width=0.125\textwidth]{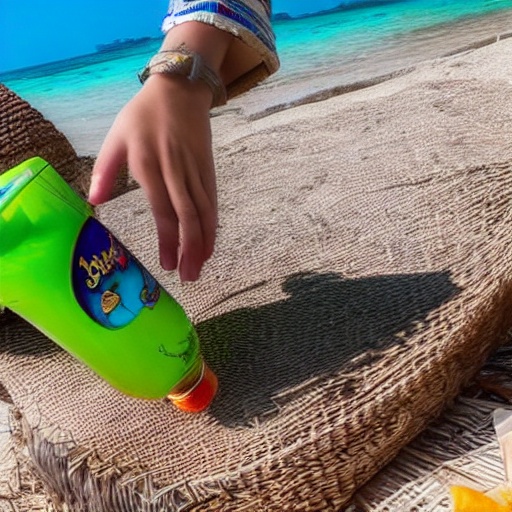}  &
         \includegraphics[width=0.125\textwidth]{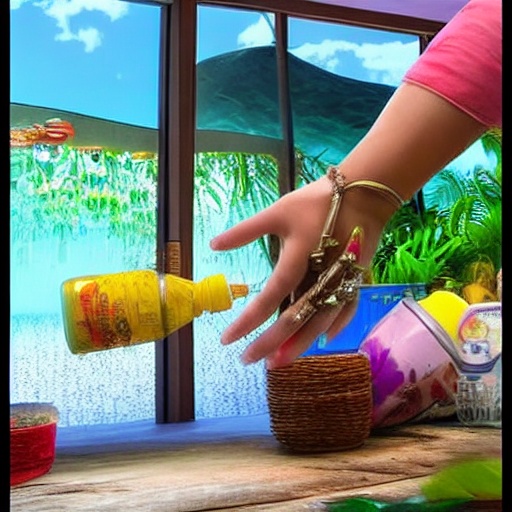} &
         \includegraphics[width=0.125\textwidth]{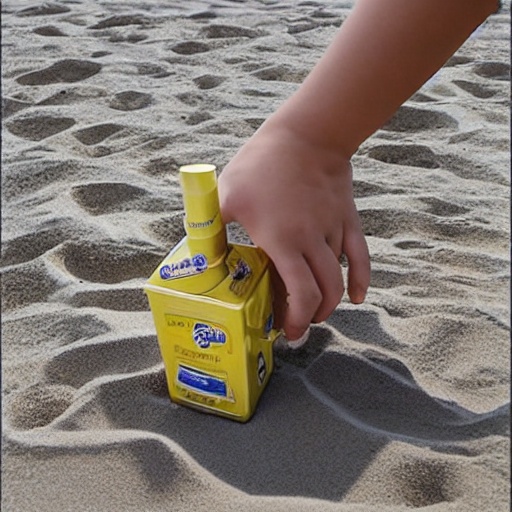}  &
         \includegraphics[width=0.125\textwidth]{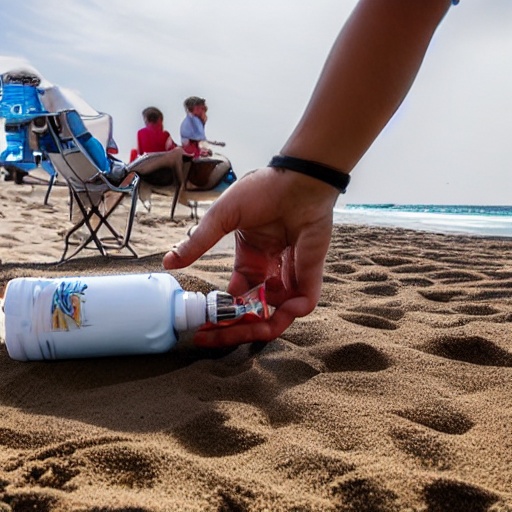} &
         \includegraphics[width=0.125\textwidth]{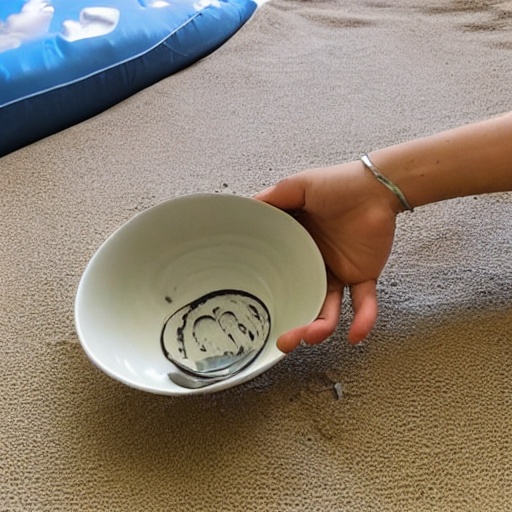} &
         \includegraphics[width=0.125\textwidth]{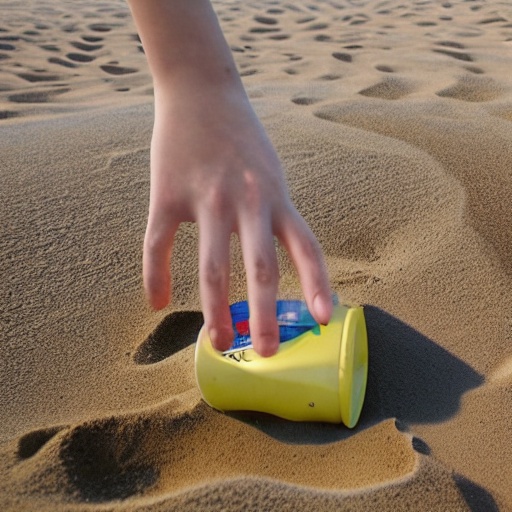}
         \\
         \\
    \multicolumn{4}{c}{Floating in the sea}&\multicolumn{4}{c}{In the Grand Canyon}\\
         \includegraphics[width=0.125\textwidth]{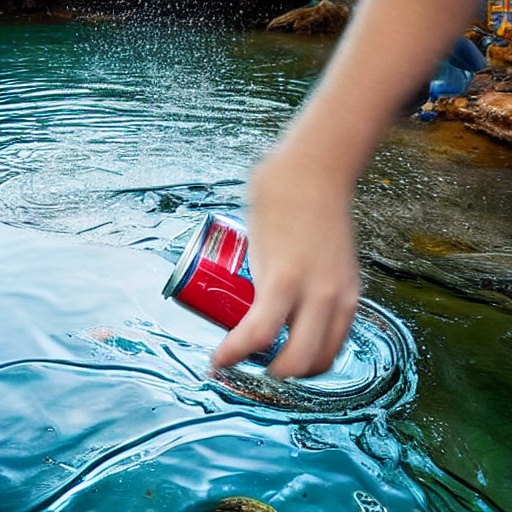} &
         \includegraphics[width=0.125\textwidth]{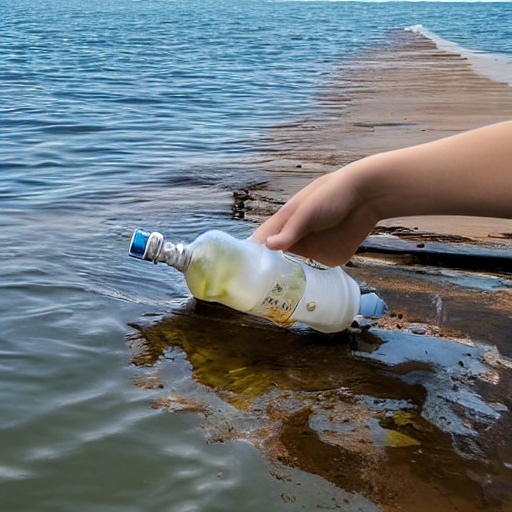}  &
         \includegraphics[width=0.125\textwidth]{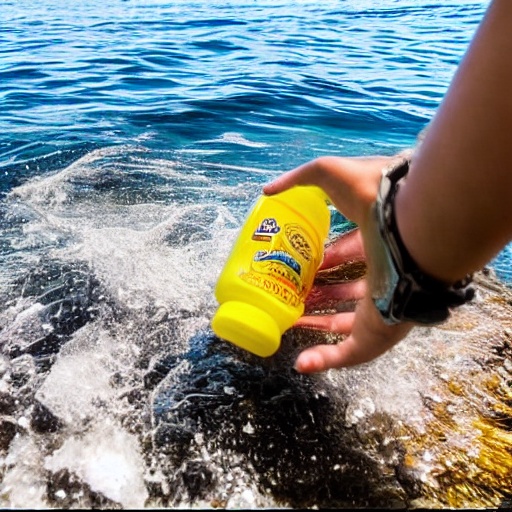}  &
         \includegraphics[width=0.125\textwidth]{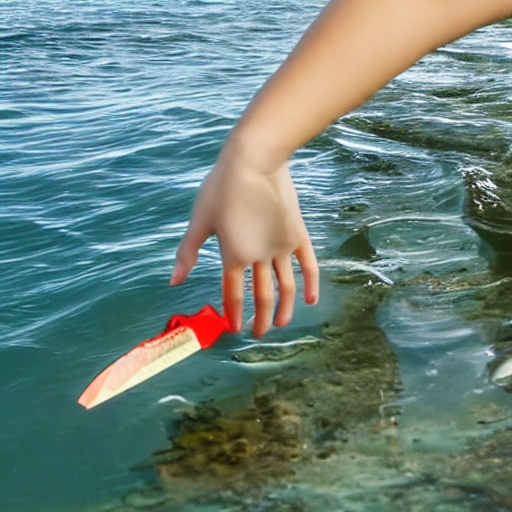} &
         \includegraphics[width=0.125\textwidth]{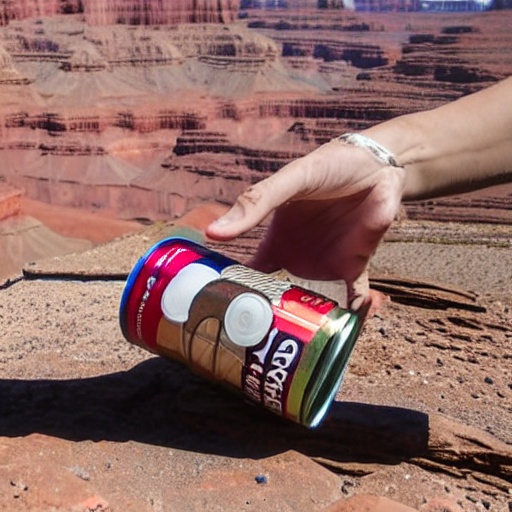}  &
         \includegraphics[width=0.125\textwidth]{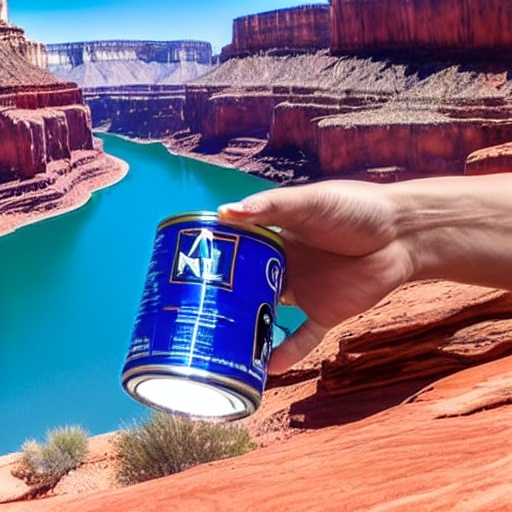} &
         \includegraphics[width=0.125\textwidth]{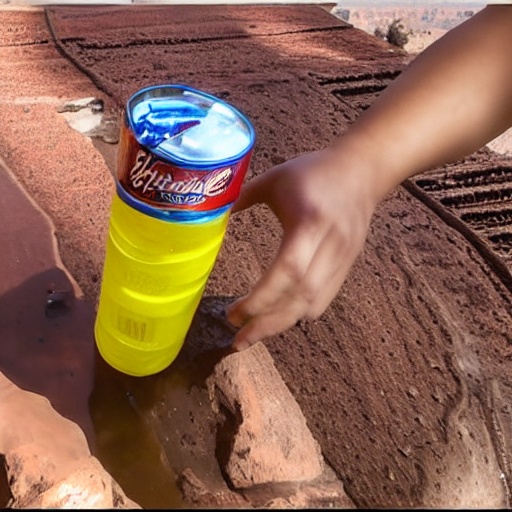} &
         \includegraphics[width=0.125\textwidth]{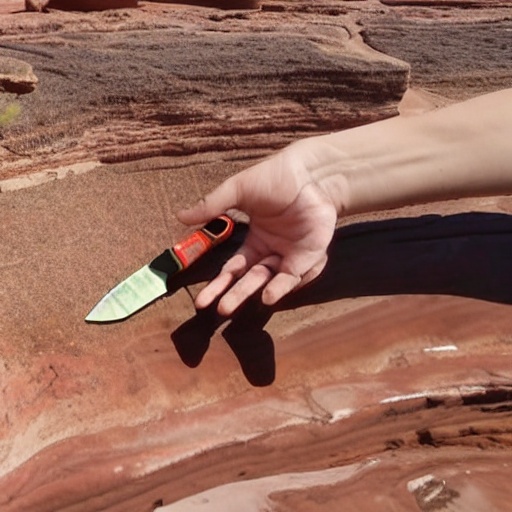}
         \\
         \\
         \multicolumn{4}{c}{In the city of Versailles}&\multicolumn{4}{c}{In an alien city on a distant planet}\\
         \includegraphics[width=0.125\textwidth]{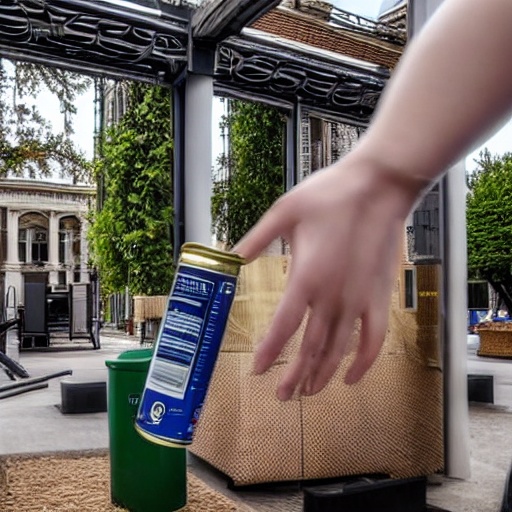} &
         \includegraphics[width=0.125\textwidth]{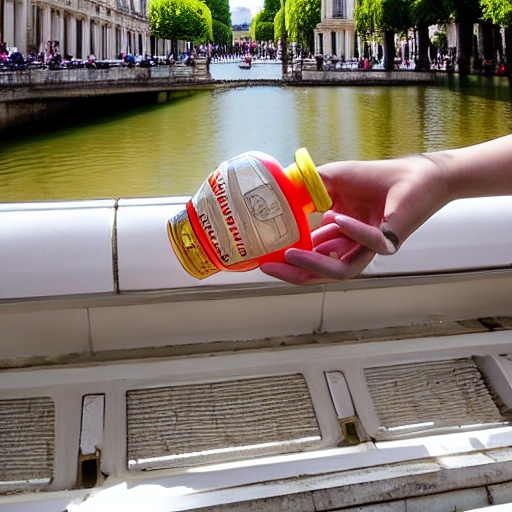}  &
         \includegraphics[width=0.125\textwidth]{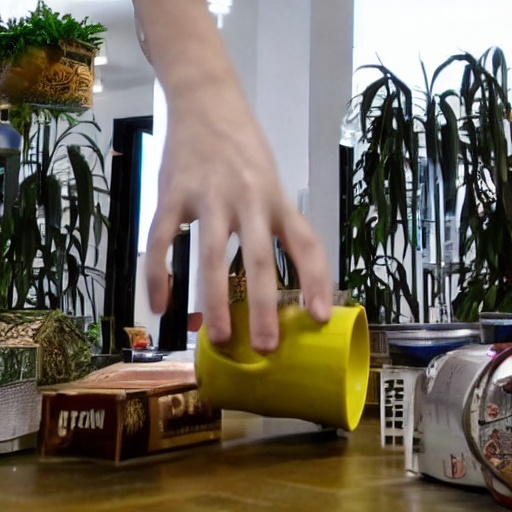}  &
         \includegraphics[width=0.125\textwidth]{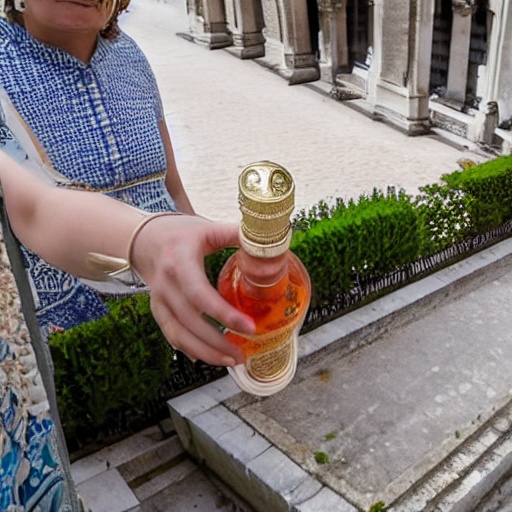} &
         \includegraphics[width=0.125\textwidth]{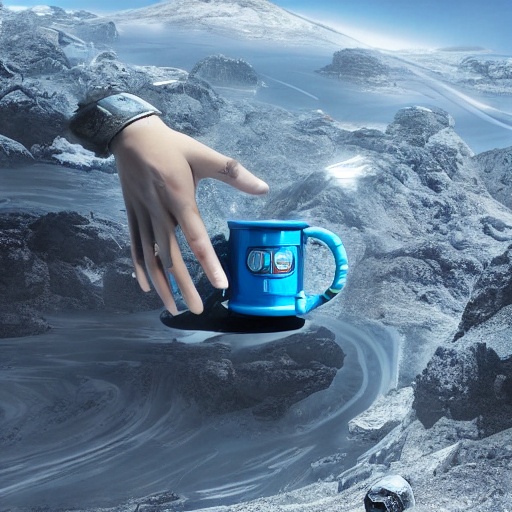}  &
         \includegraphics[width=0.125\textwidth]{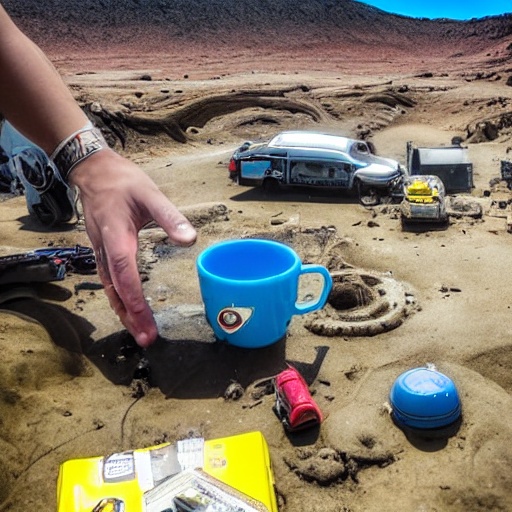} &
         \includegraphics[width=0.125\textwidth]{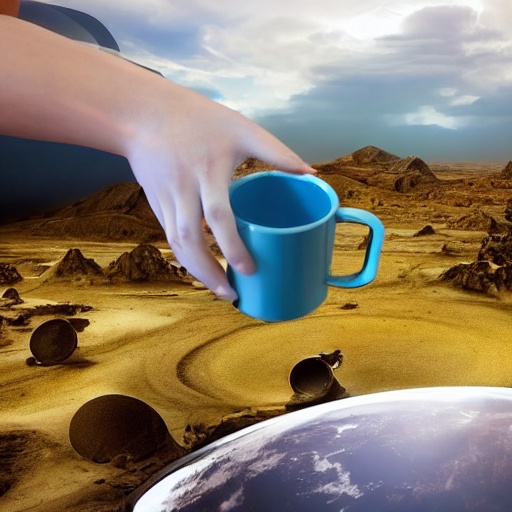} &
         \includegraphics[width=0.125\textwidth]{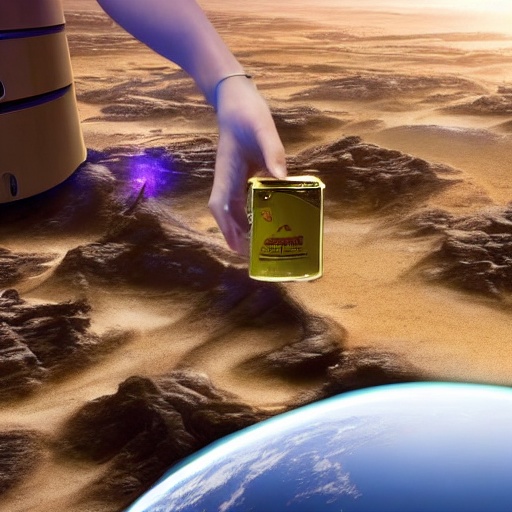} \\
         \\
         \multicolumn{4}{c}{In the underwater city of Atlantis}&\multicolumn{4}{c}{In a Japanese rock garden}\\

        \includegraphics[width=0.125\textwidth]{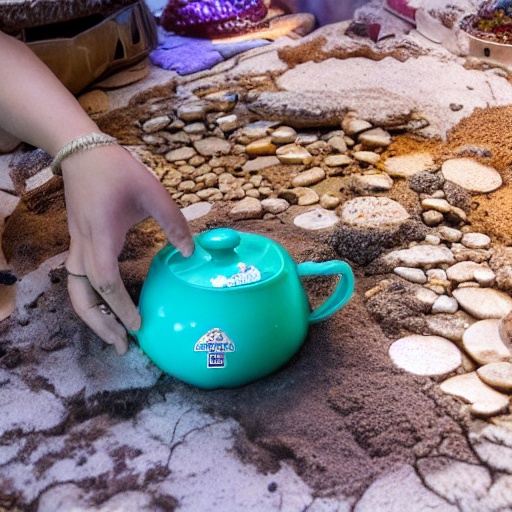} &
         \includegraphics[width=0.125\textwidth]{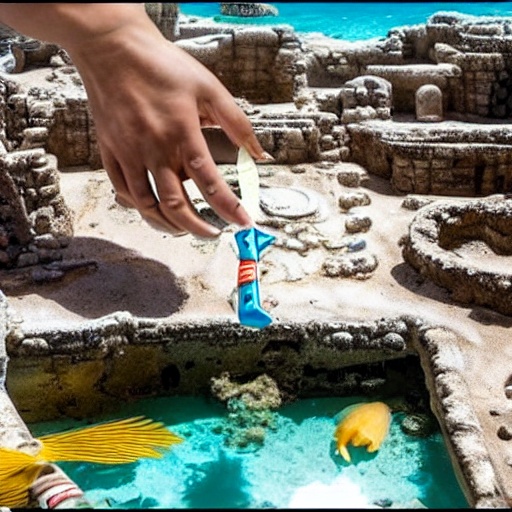}  &
         \includegraphics[width=0.125\textwidth]{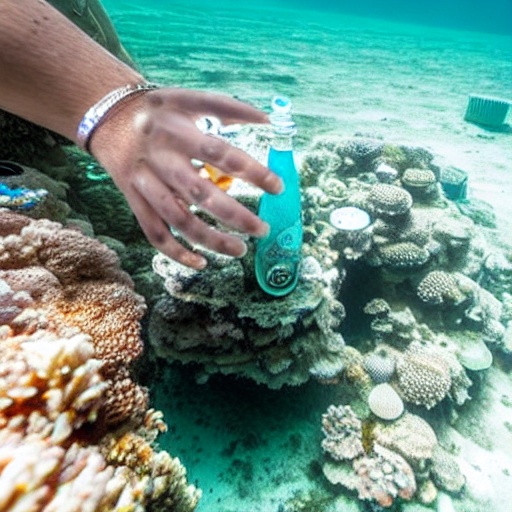}  &
         \includegraphics[width=0.125\textwidth]{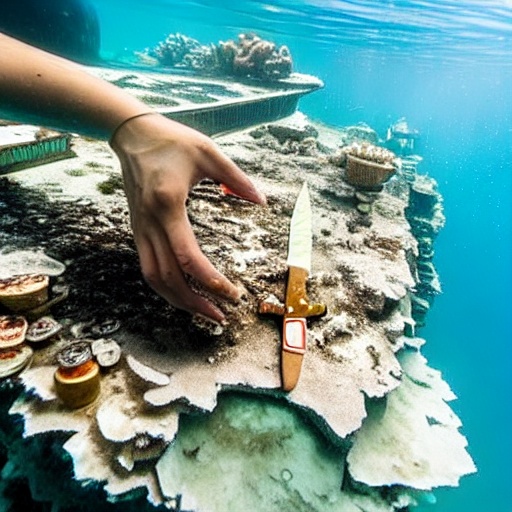} &
         \includegraphics[width=0.125\textwidth]{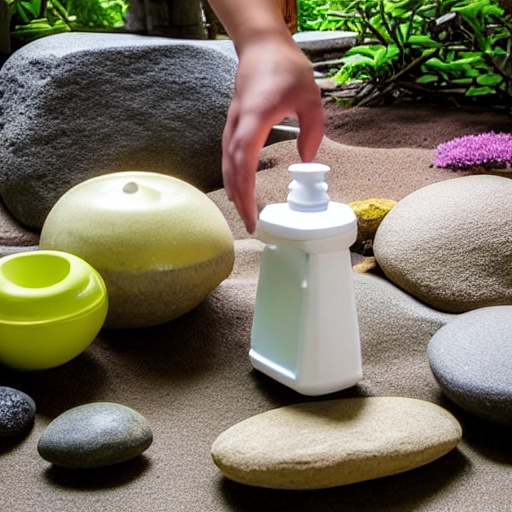}  &
         \includegraphics[width=0.125\textwidth]{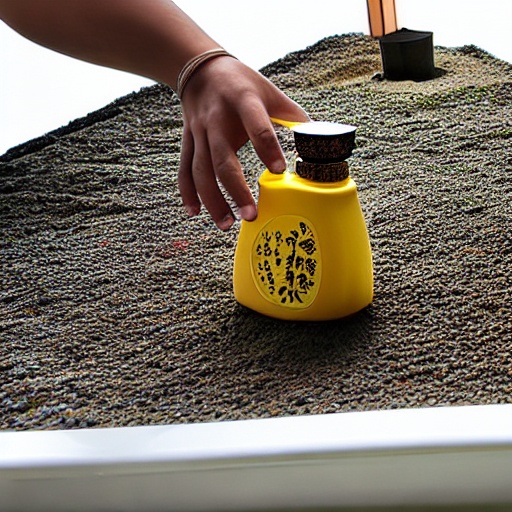} &
         \includegraphics[width=0.125\textwidth]{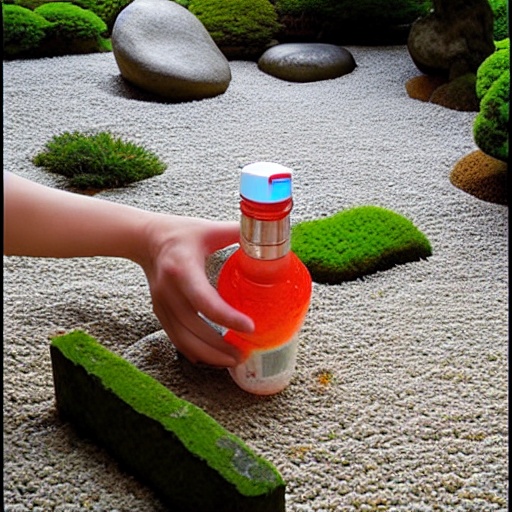} &
         \includegraphics[width=0.125\textwidth]{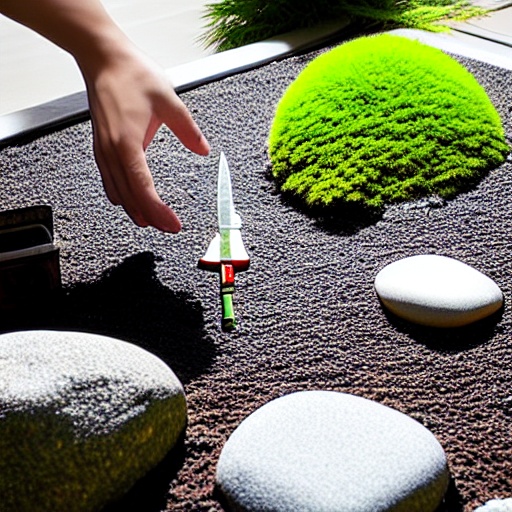} \\
         \\
        \multicolumn{4}{c}{In a lavender field}&\multicolumn{4}{c}{In neon-lit cyberpunk nightclub}\\
         \includegraphics[width=0.125\textwidth]{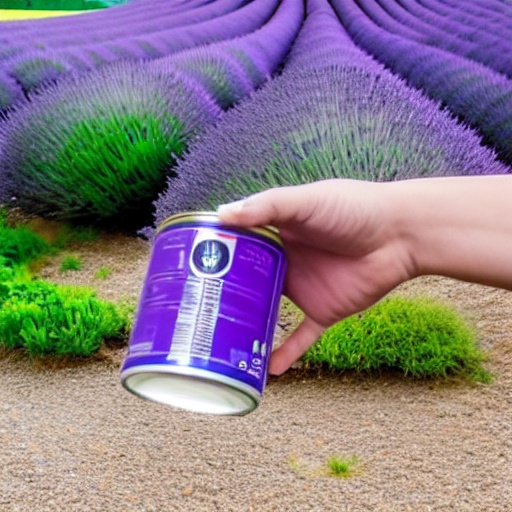} &
         \includegraphics[width=0.125\textwidth]{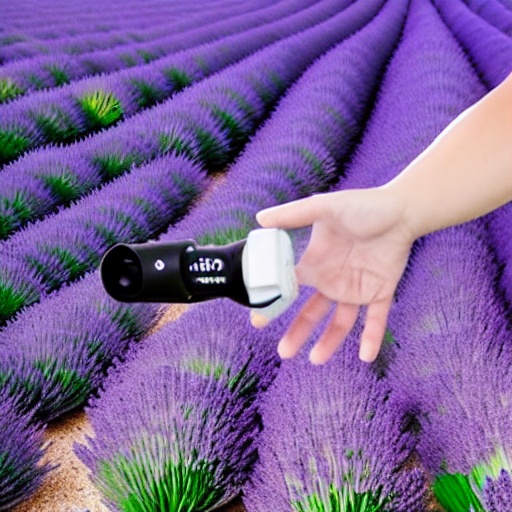}  &
         \includegraphics[width=0.125\textwidth]{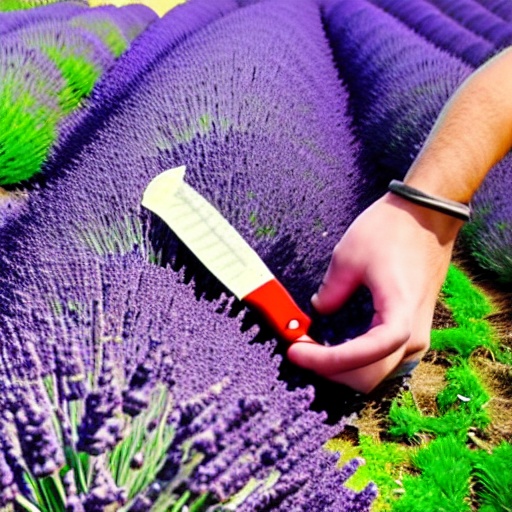}  &
         \includegraphics[width=0.125\textwidth]{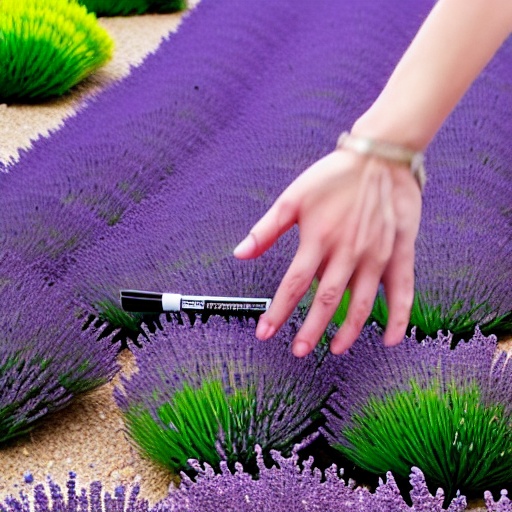} &
         \includegraphics[width=0.125\textwidth]{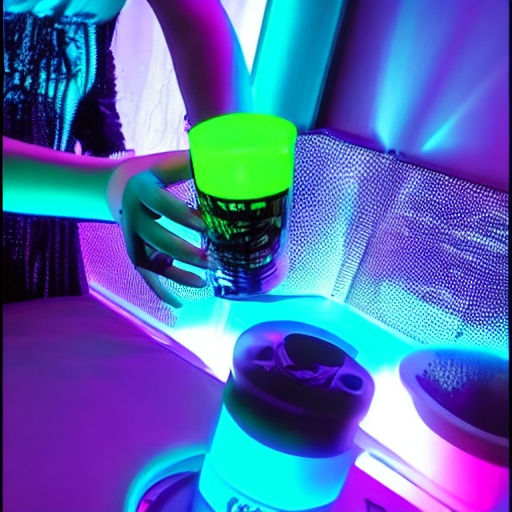}  &
         \includegraphics[width=0.125\textwidth]{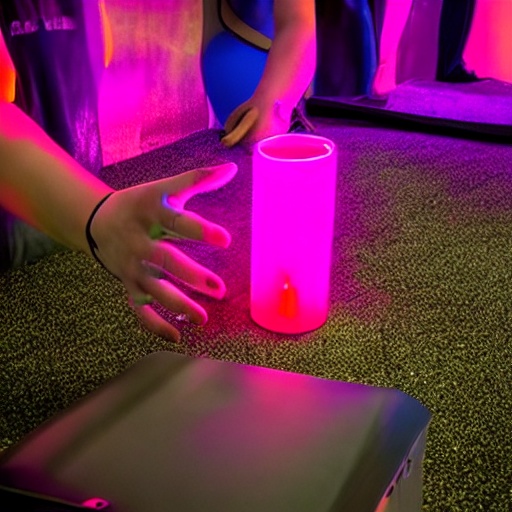} &
         \includegraphics[width=0.125\textwidth]{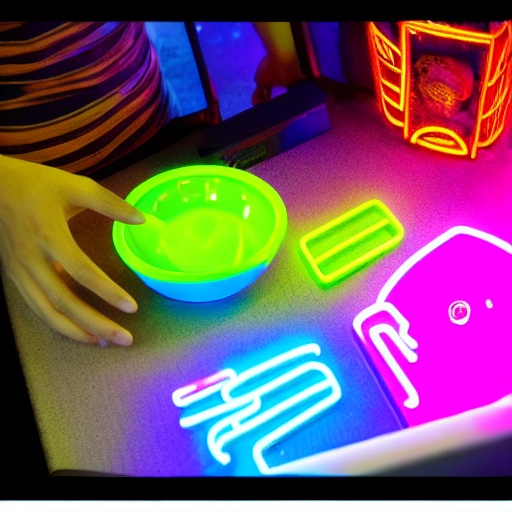} &
         \includegraphics[width=0.125\textwidth]{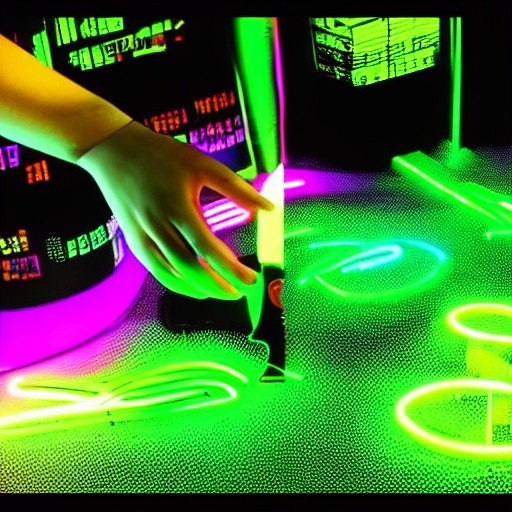} \\
         \\
        \multicolumn{4}{c}{In the snow}&\multicolumn{4}{c}{In bamboo forest}\\
         \includegraphics[width=0.125\textwidth]{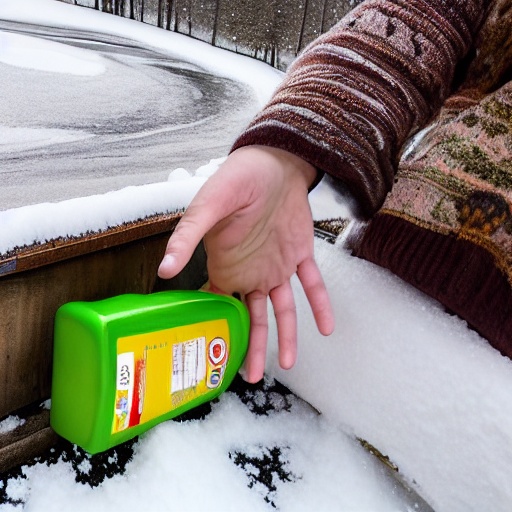} &
         \includegraphics[width=0.125\textwidth]{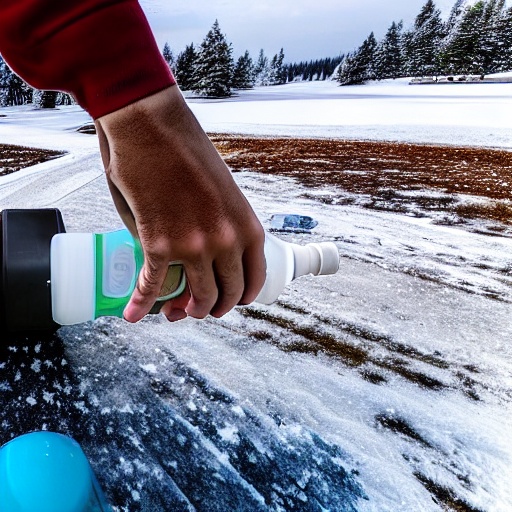}  &
         \includegraphics[width=0.125\textwidth]{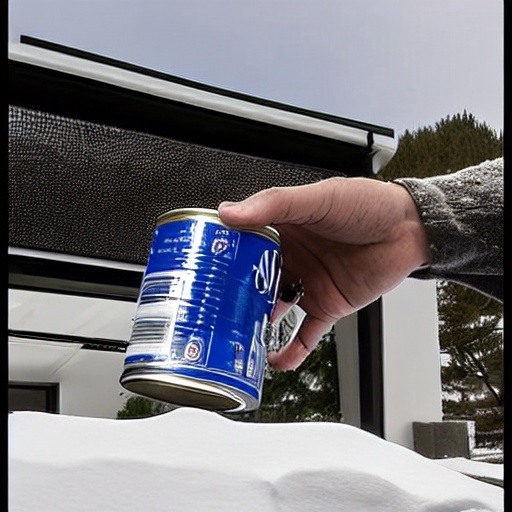}  &
         \includegraphics[width=0.125\textwidth]{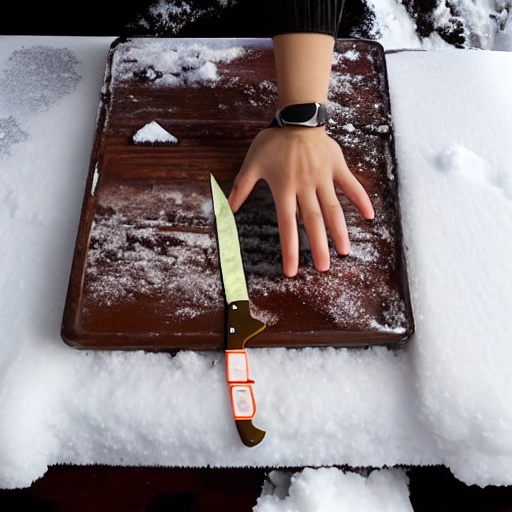} &
         \includegraphics[width=0.125\textwidth]{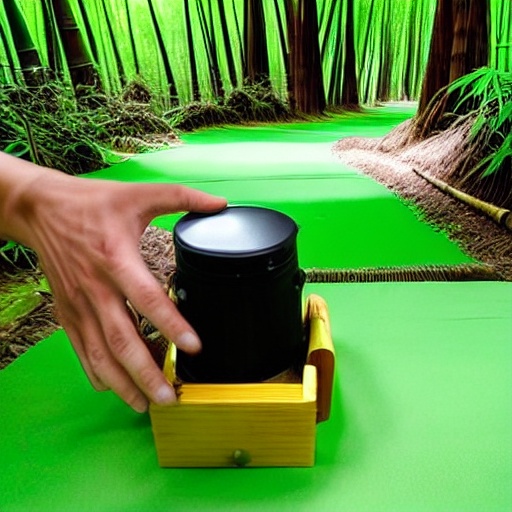}  &
         \includegraphics[width=0.125\textwidth]{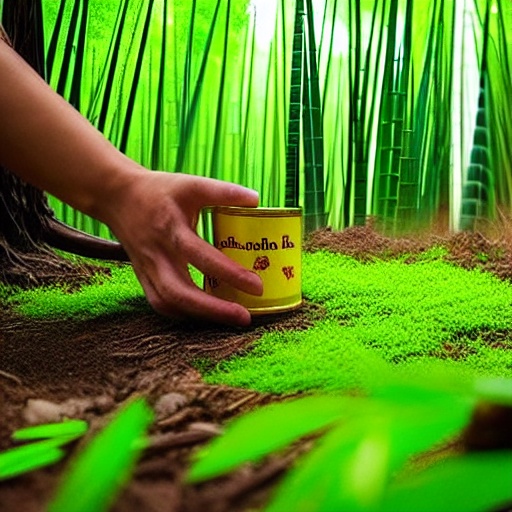} &
         \includegraphics[width=0.125\textwidth]{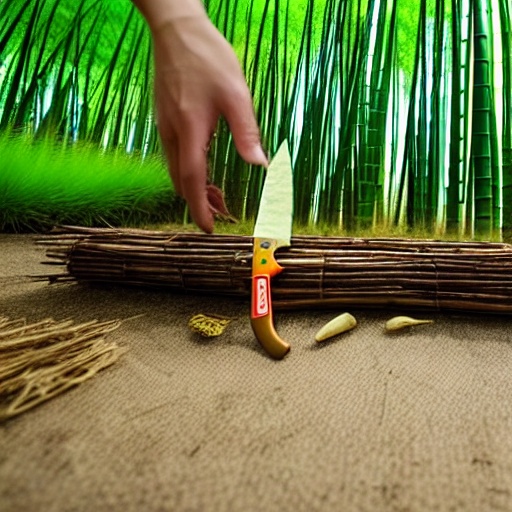} &
         \includegraphics[width=0.125\textwidth]{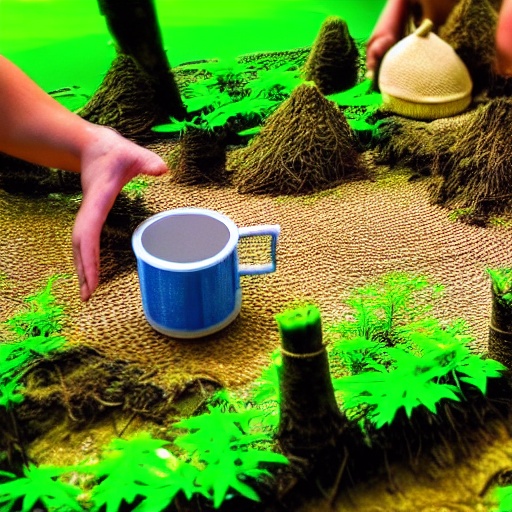} \\
         \\
    \end{tabular}
    }
    \vspace{-3mm}
    \caption{Generated images with more text prompts ranging from daily landscape to virtual scene.}
    \vspace{-4mm}
    \label{fig:sup_back}
\end{figure*}

\section{Background Control}
\label{sec:back}
 In this section, we present more generated hand-object-interaction images with various background prompts, from everyday scenarios to special contexts. We refer readers to Figure \ref{fig:sup_back}. 

\begin{figure*}
\centering
\scalebox{0.98}{
    \setlength\tabcolsep{0.5pt}
    \renewcommand{\arraystretch}{0.25}
    \begin{tabular}{cccccccc}
         \includegraphics[width=0.125\textwidth]{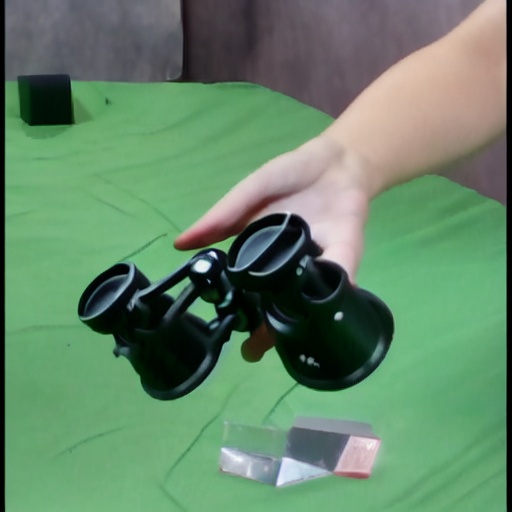} &
         \includegraphics[width=0.125\textwidth]{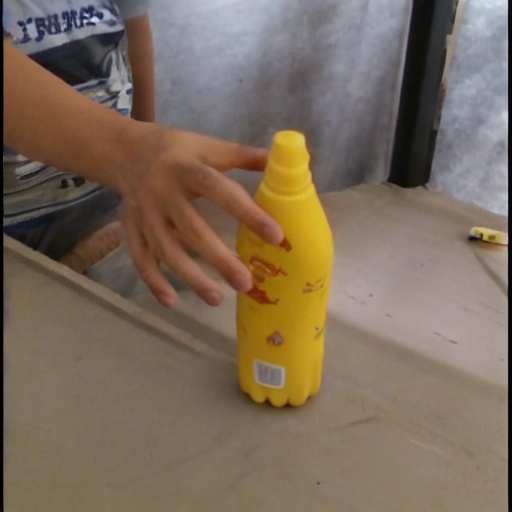}  &
         \includegraphics[width=0.125\textwidth]{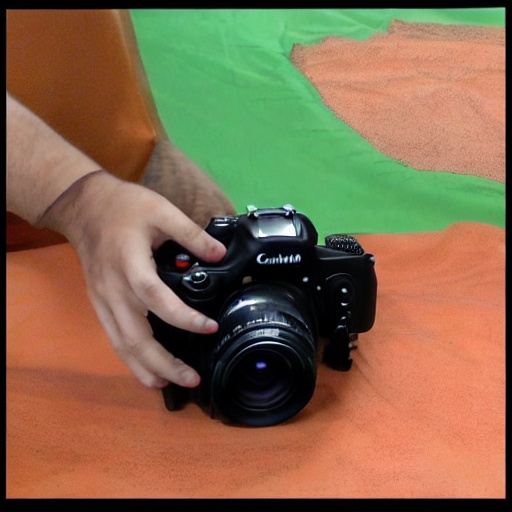}  &
         \includegraphics[width=0.125\textwidth]{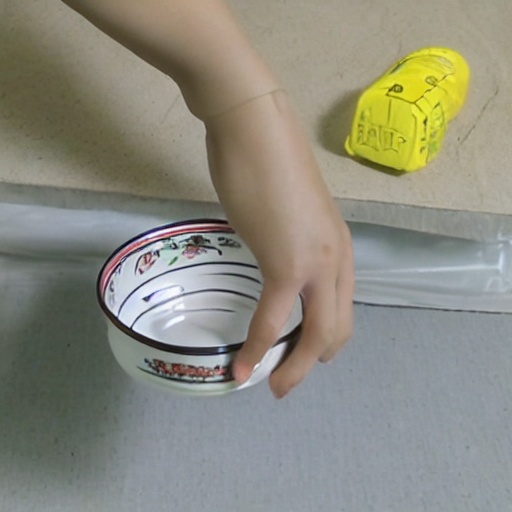} &
         \includegraphics[width=0.125\textwidth]{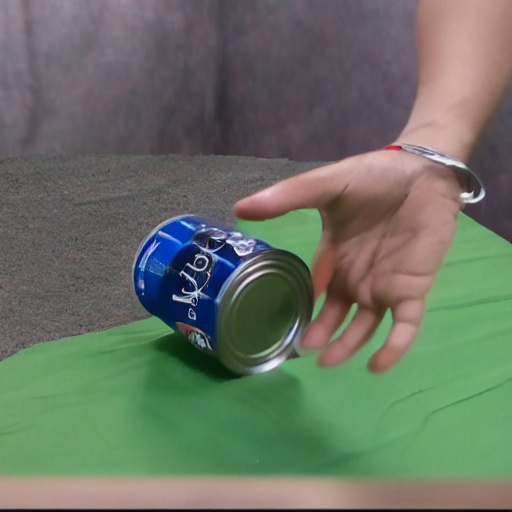}  &
         \includegraphics[width=0.125\textwidth]{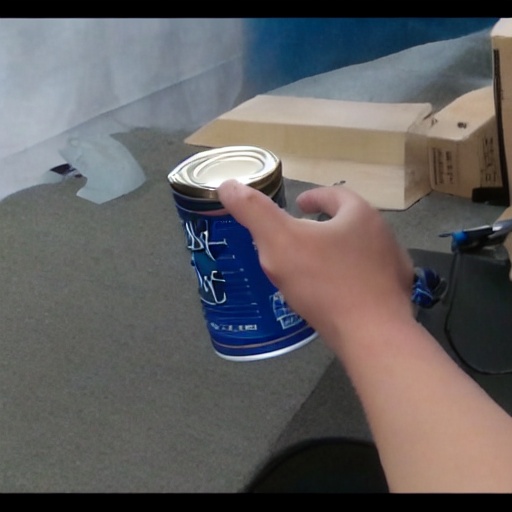} &
         \includegraphics[width=0.125\textwidth]{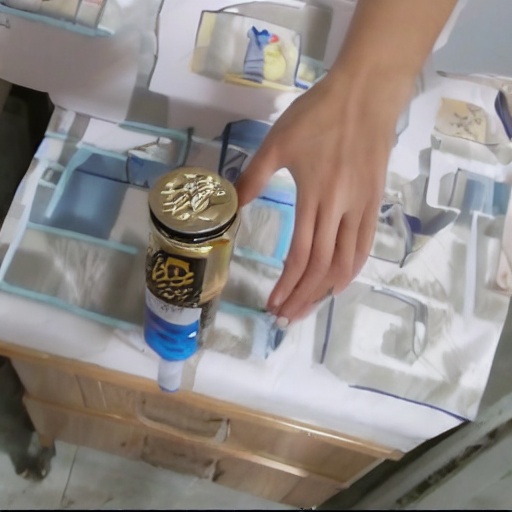} &
         \includegraphics[width=0.125\textwidth]{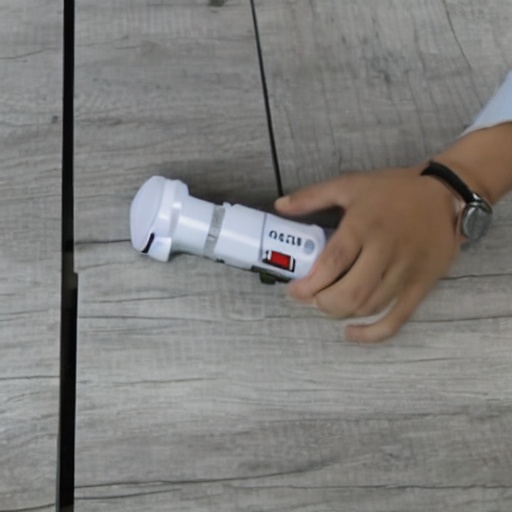}
         \\
         \includegraphics[width=0.125\textwidth]{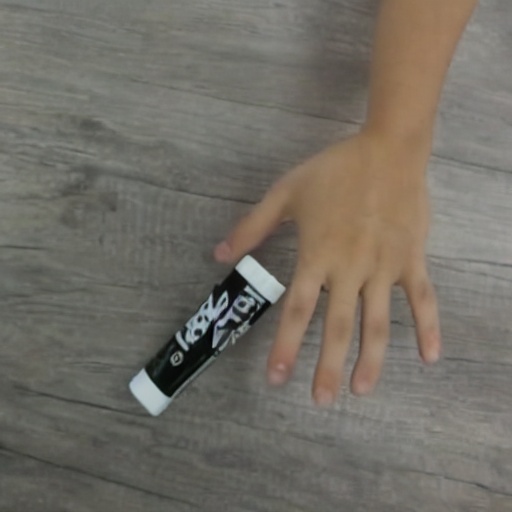} &
         \includegraphics[width=0.125\textwidth]{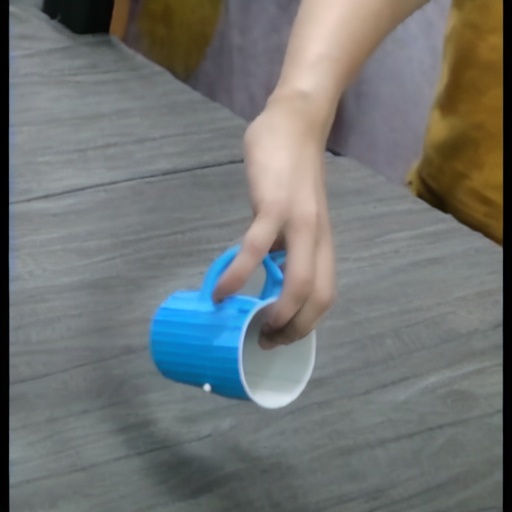}  &
         \includegraphics[width=0.125\textwidth]{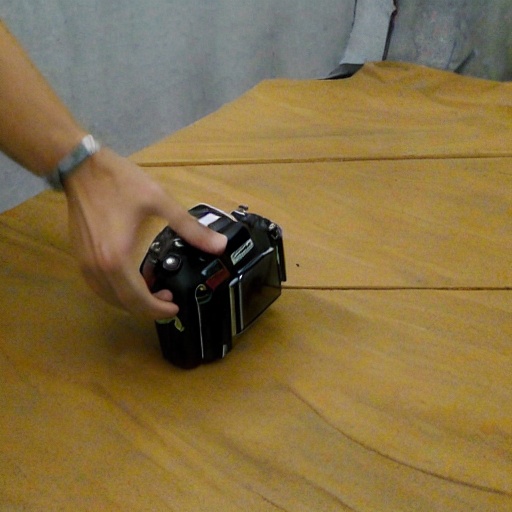}  &
         \includegraphics[width=0.125\textwidth]{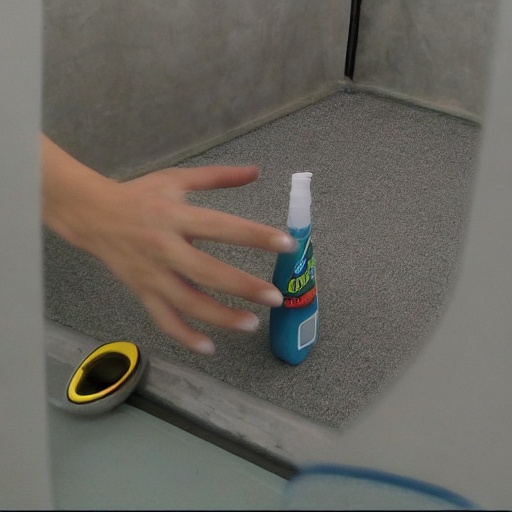} &
         \includegraphics[width=0.125\textwidth]{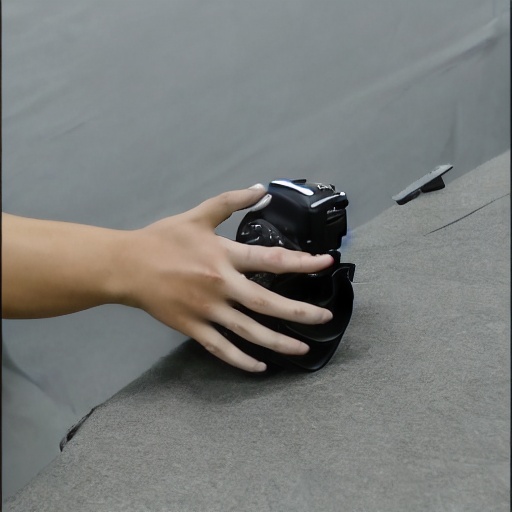}  &
         \includegraphics[width=0.125\textwidth]{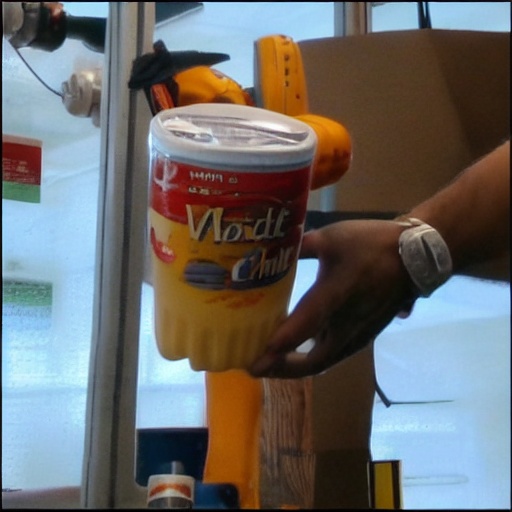} &
         \includegraphics[width=0.125\textwidth]{images/sup/angle/bowl_s154_0_2_11.jpg} &
         \includegraphics[width=0.125\textwidth]{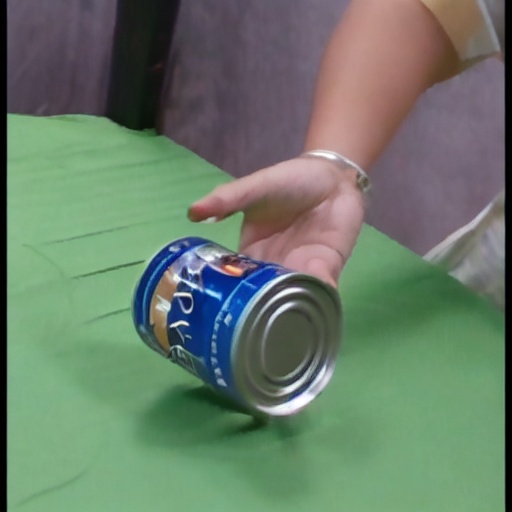} \\
         
         \includegraphics[width=0.125\textwidth]{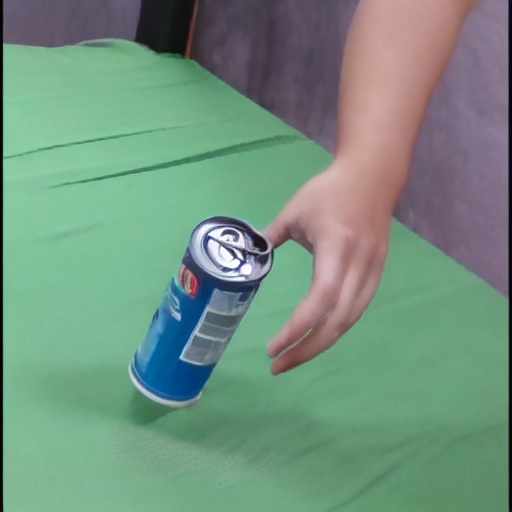} &
         \includegraphics[width=0.125\textwidth]{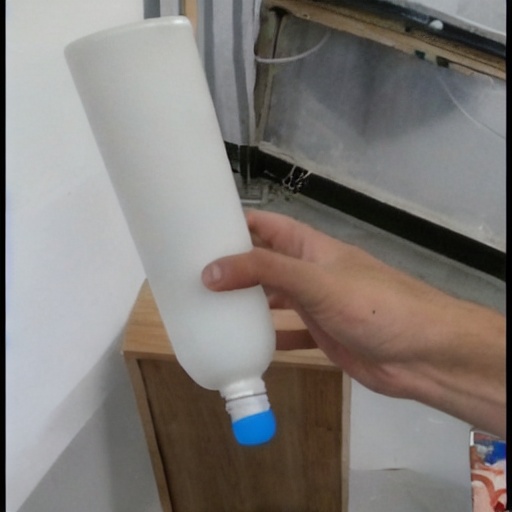}  &
         \includegraphics[width=0.125\textwidth]{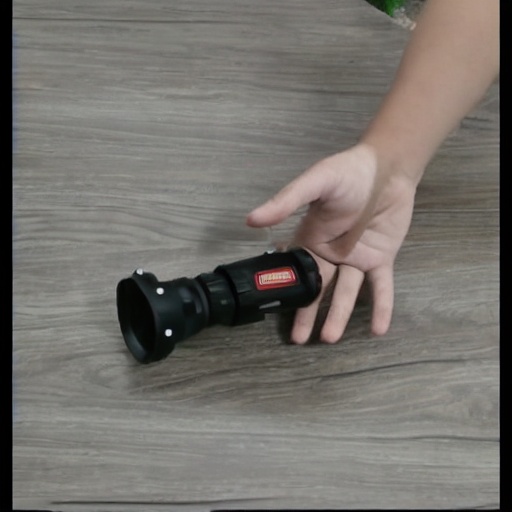}  &
         \includegraphics[width=0.125\textwidth]{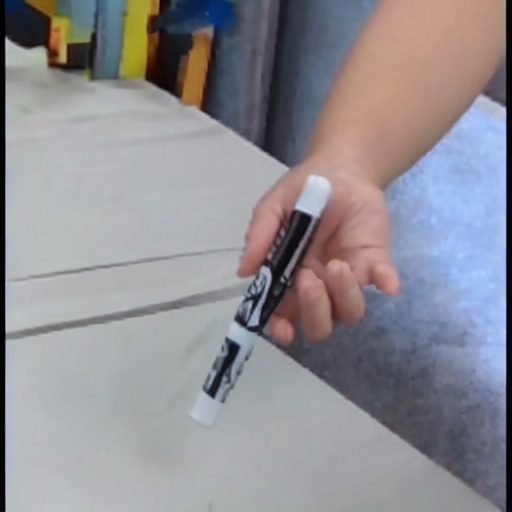} &
         \includegraphics[width=0.125\textwidth]{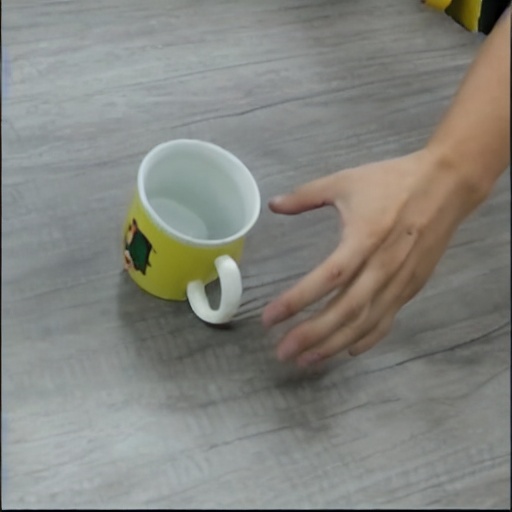}  &
         \includegraphics[width=0.125\textwidth]{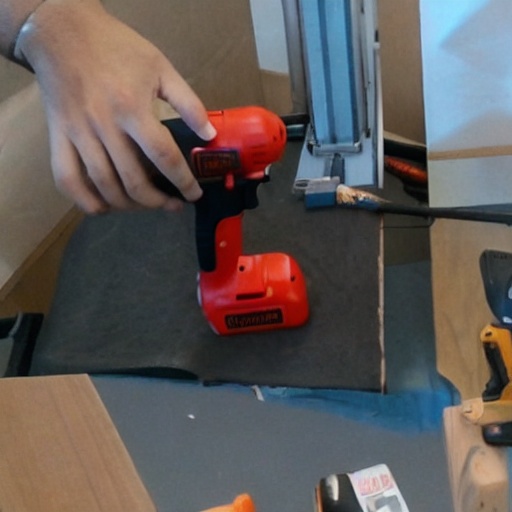} &
         \includegraphics[width=0.125\textwidth]{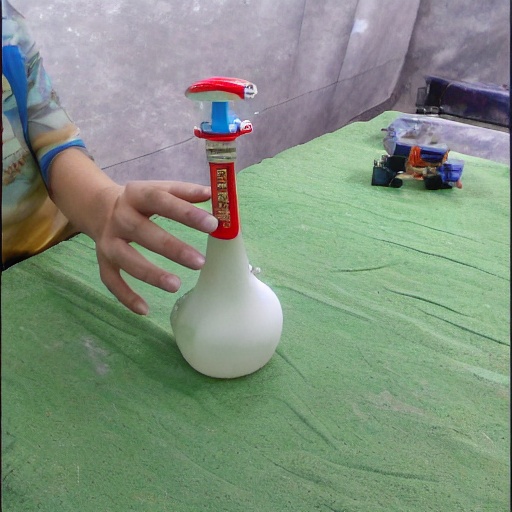} &
         \includegraphics[width=0.125\textwidth]{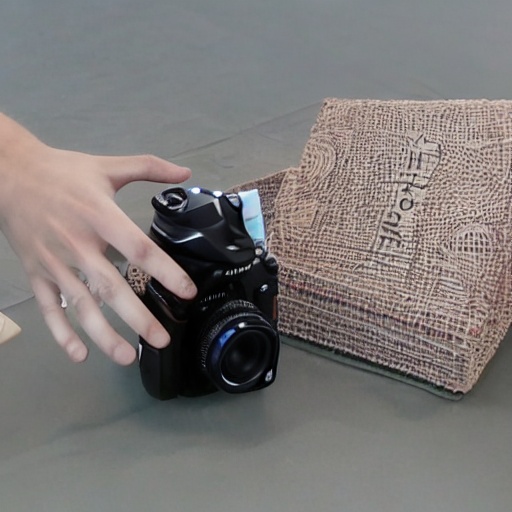} \\

         \includegraphics[width=0.125\textwidth]{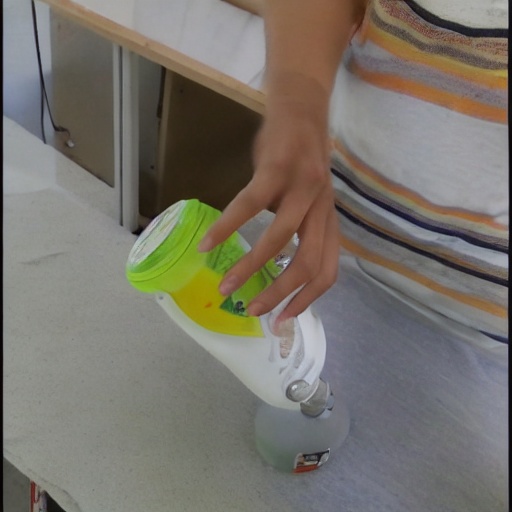} &
         \includegraphics[width=0.125\textwidth]{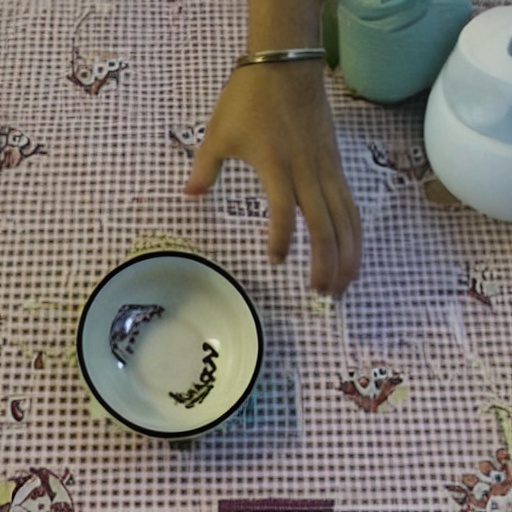}  &
         \includegraphics[width=0.125\textwidth]{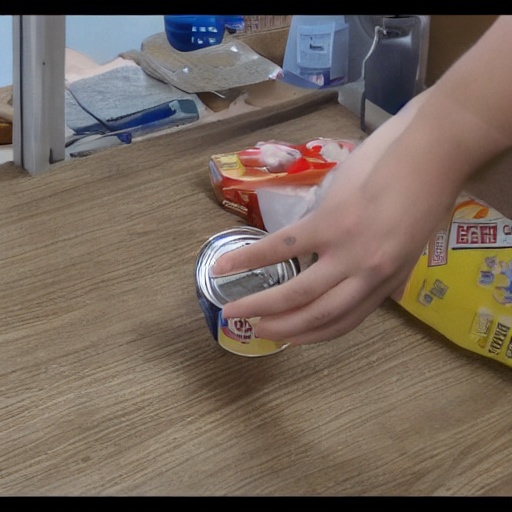}  &
         \includegraphics[width=0.125\textwidth]{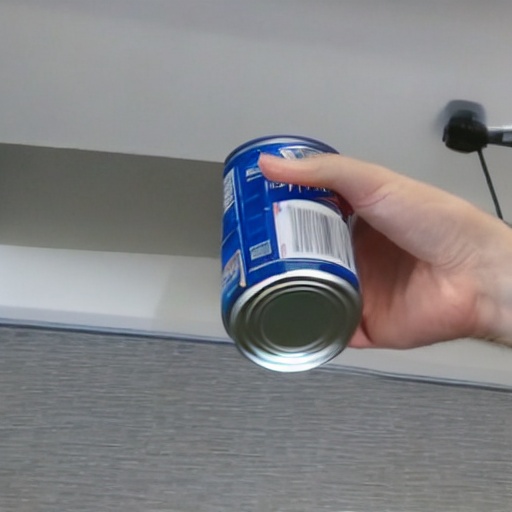} &
         \includegraphics[width=0.125\textwidth]{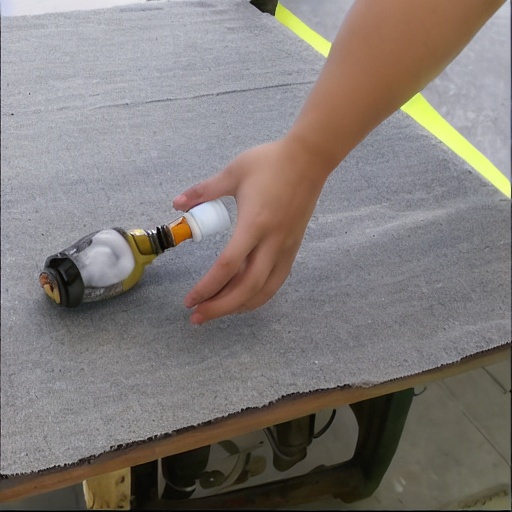}  &
         \includegraphics[width=0.125\textwidth]{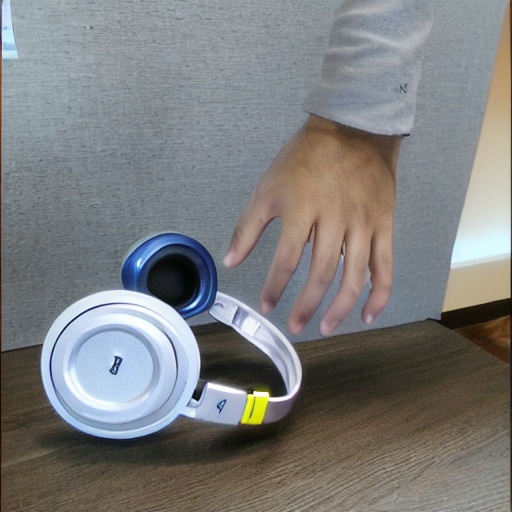} &
         \includegraphics[width=0.125\textwidth]{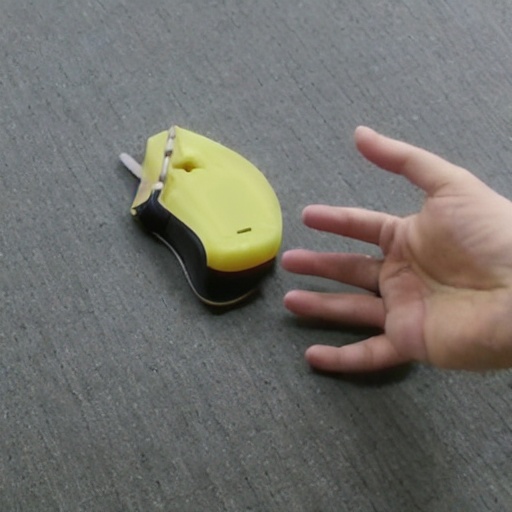} &
         \includegraphics[width=0.125\textwidth]{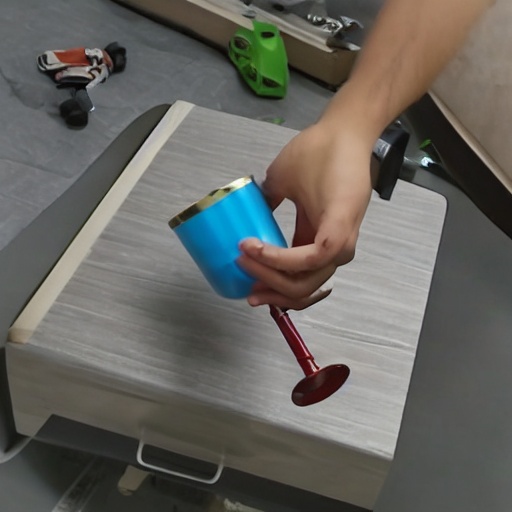} \\

         \includegraphics[width=0.125\textwidth]{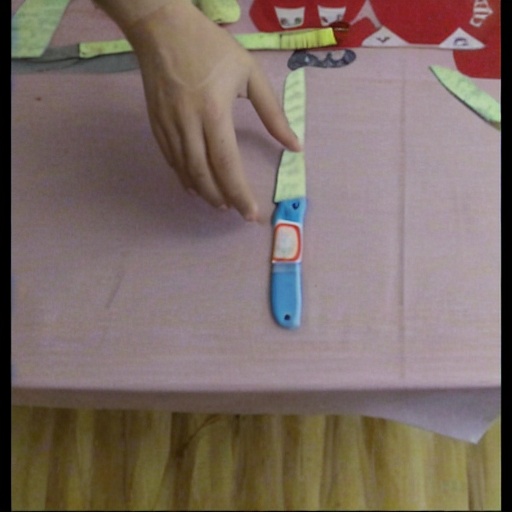} &
         \includegraphics[width=0.125\textwidth]{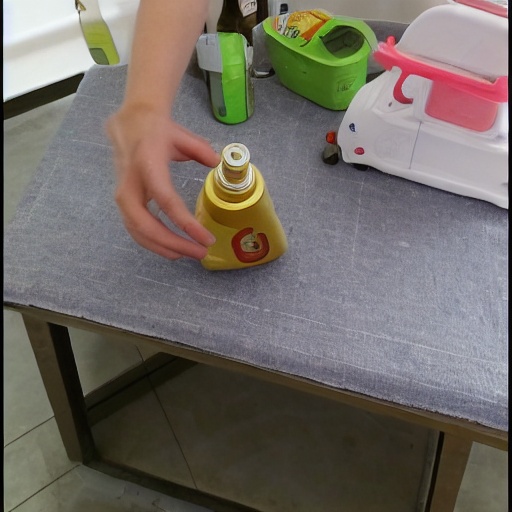}  &
         \includegraphics[width=0.125\textwidth]{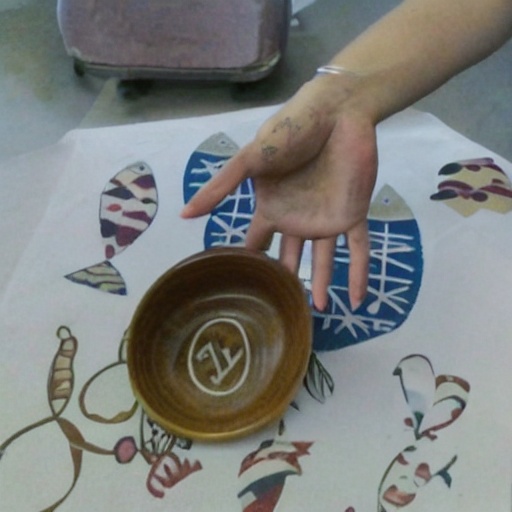}  &
         \includegraphics[width=0.125\textwidth]{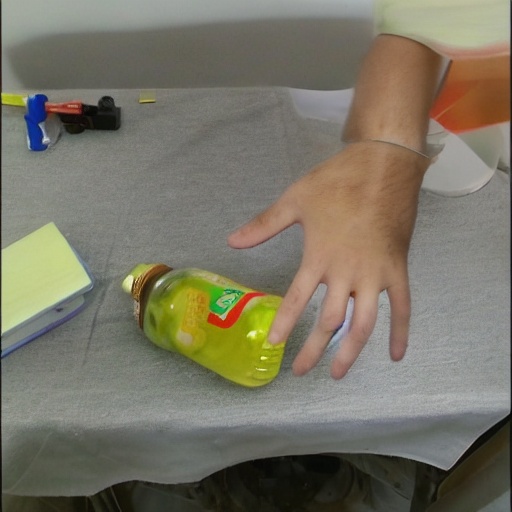} &
         \includegraphics[width=0.125\textwidth]{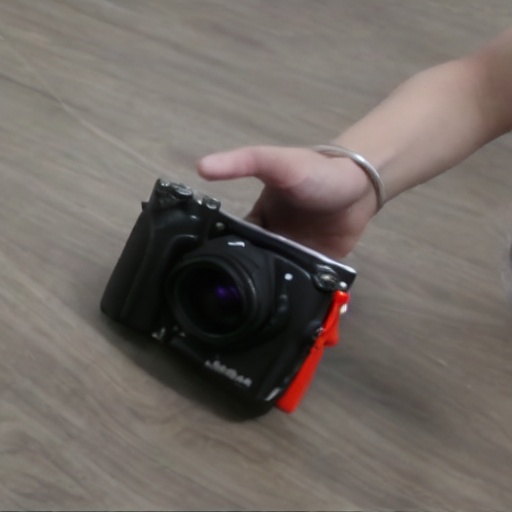}  &
         \includegraphics[width=0.125\textwidth]{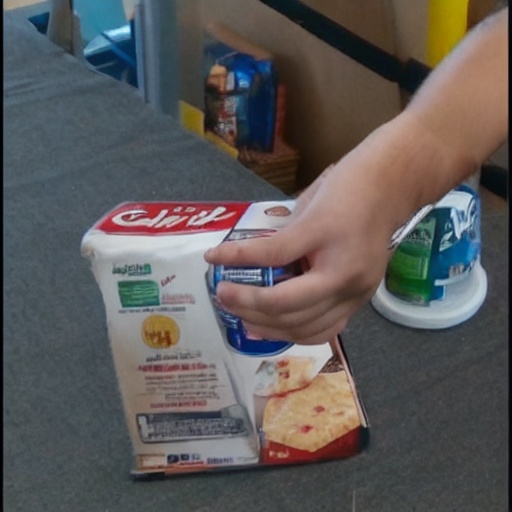} &
         \includegraphics[width=0.125\textwidth]{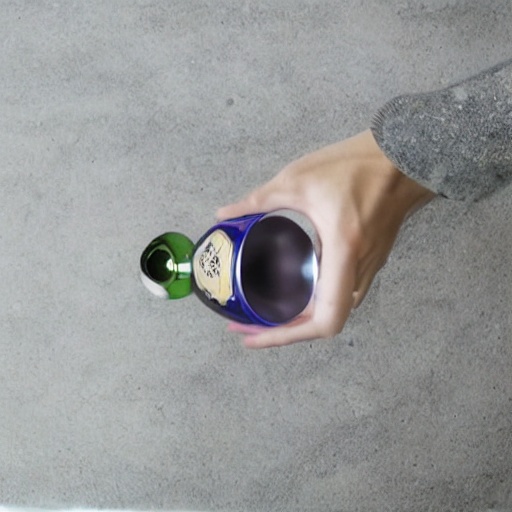} &
         \includegraphics[width=0.125\textwidth]{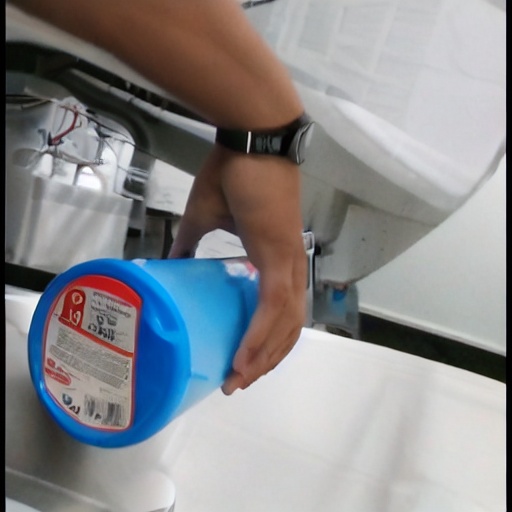} \\

         \includegraphics[width=0.125\textwidth]{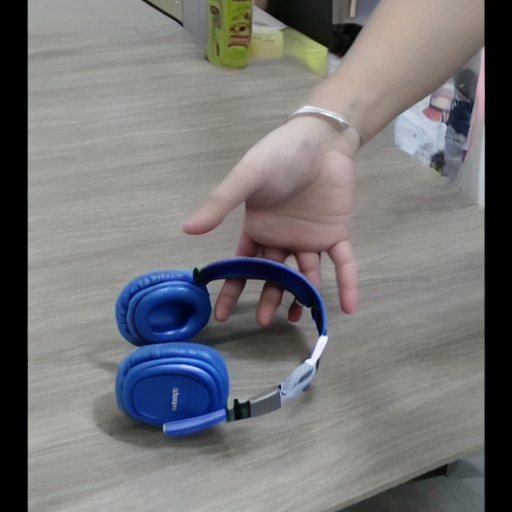} &
         \includegraphics[width=0.125\textwidth]{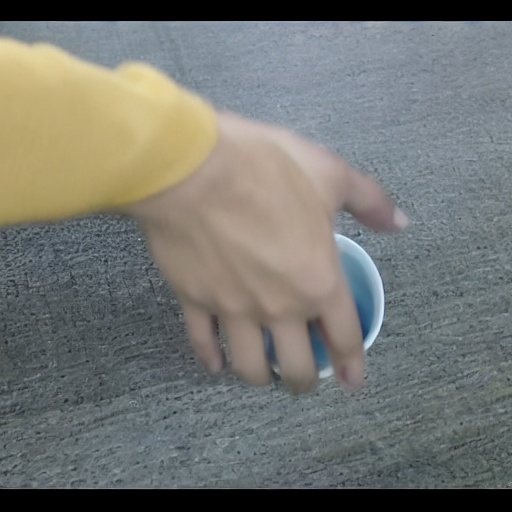}  &
         \includegraphics[width=0.125\textwidth]{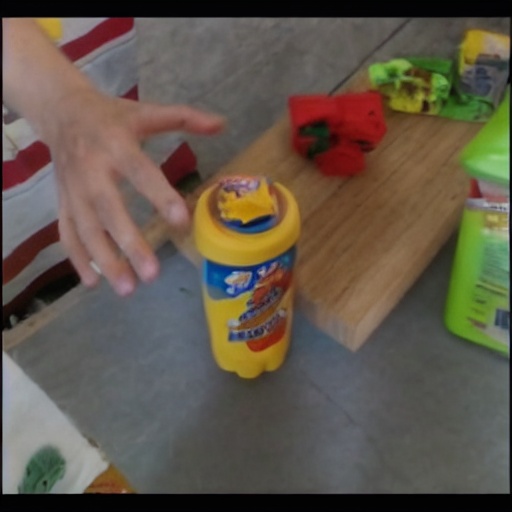}  &
         \includegraphics[width=0.125\textwidth]{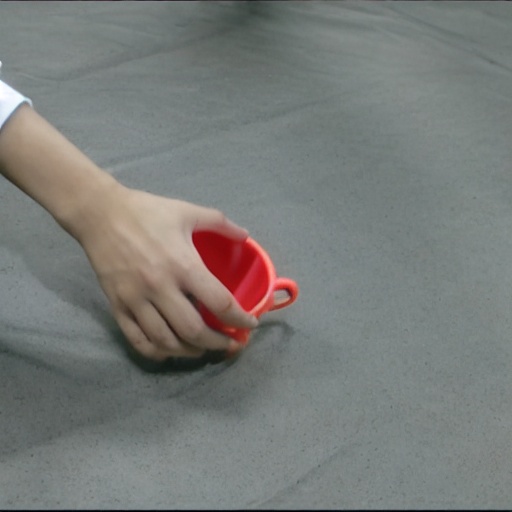} &
         \includegraphics[width=0.125\textwidth]{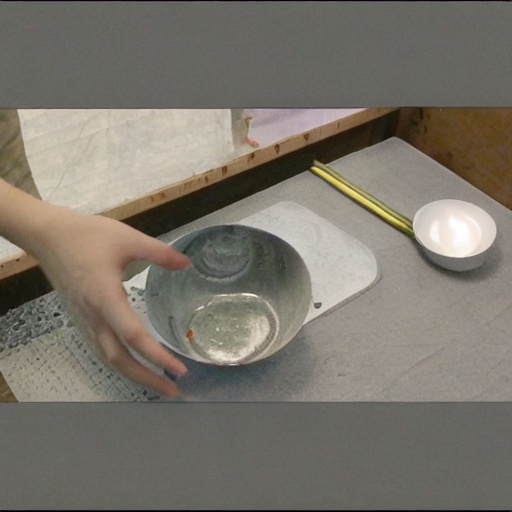}  &
         \includegraphics[width=0.125\textwidth]{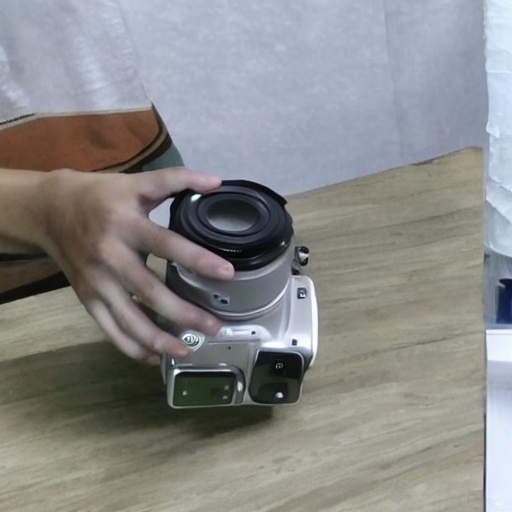} &
         \includegraphics[width=0.125\textwidth]{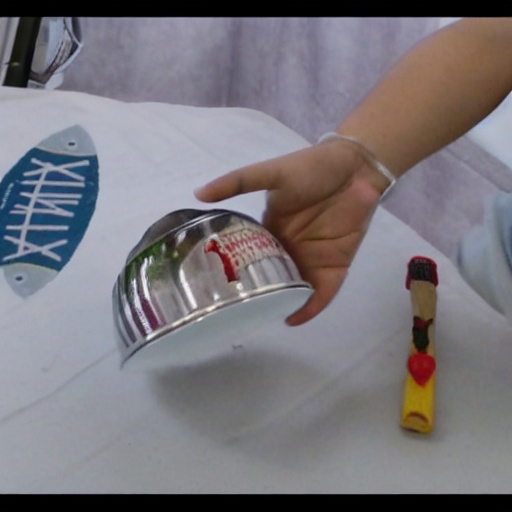} &
         \includegraphics[width=0.125\textwidth]{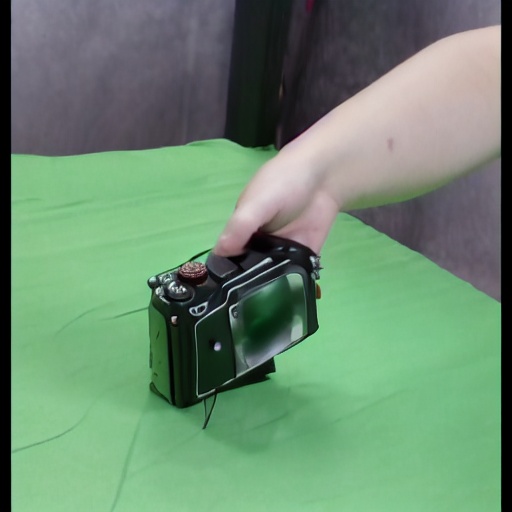} \\

         \includegraphics[width=0.125\textwidth]{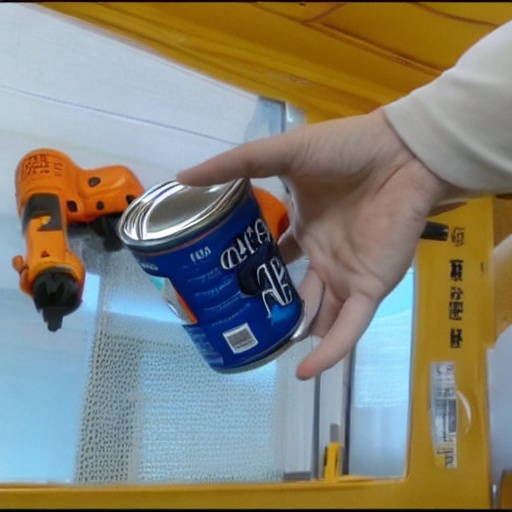} &
         \includegraphics[width=0.125\textwidth]{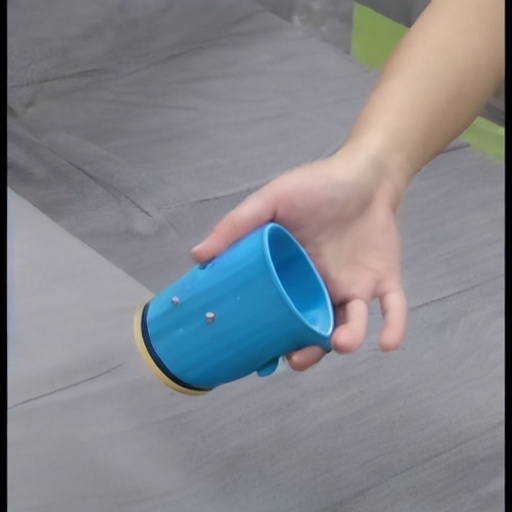}  &
         \includegraphics[width=0.125\textwidth]{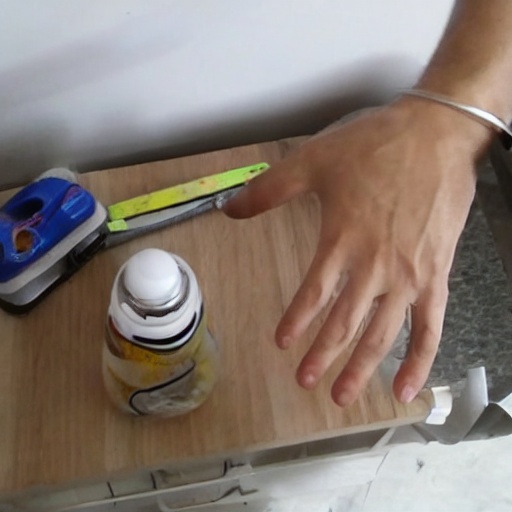}  &
         \includegraphics[width=0.125\textwidth]{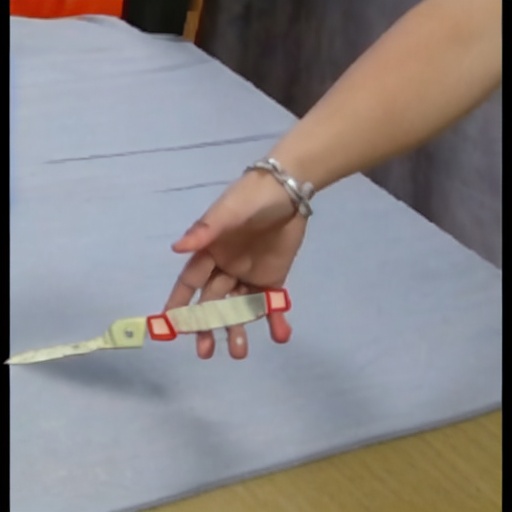} &
         \includegraphics[width=0.125\textwidth]{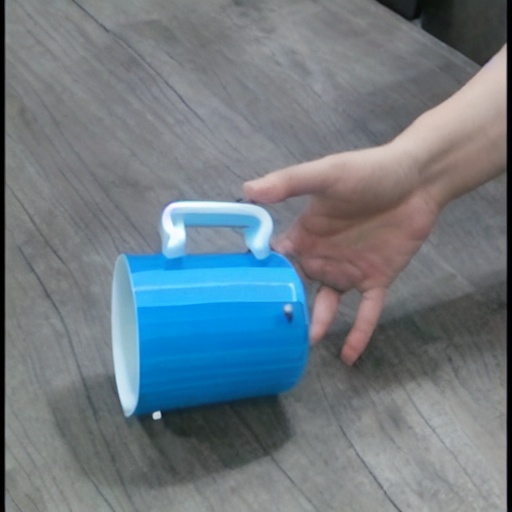}  &
         \includegraphics[width=0.125\textwidth]{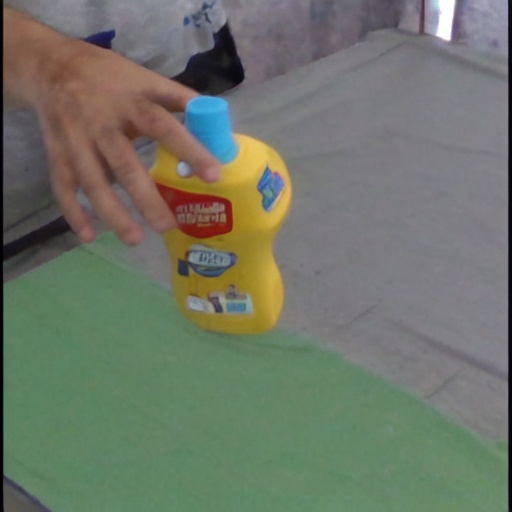} &
         \includegraphics[width=0.125\textwidth]{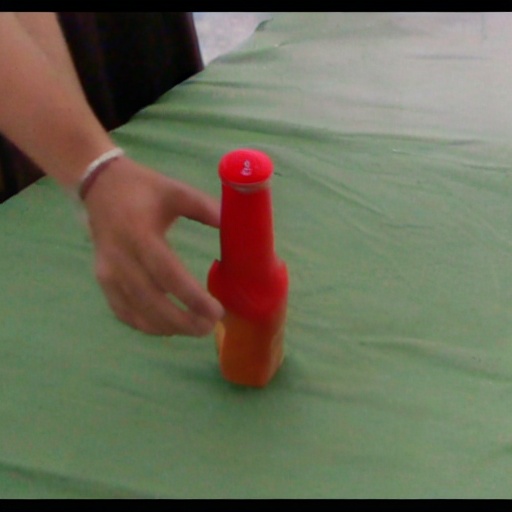} &
         \includegraphics[width=0.125\textwidth]{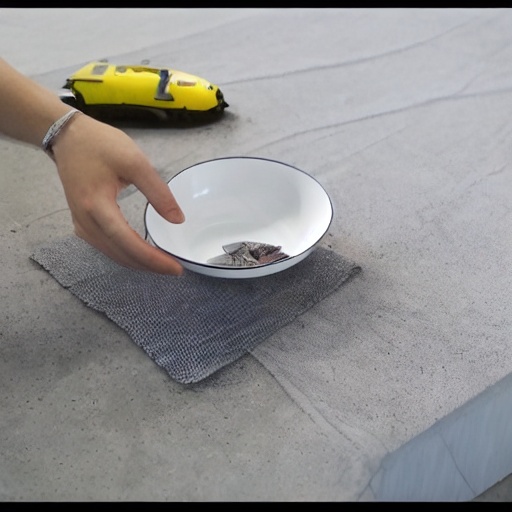} \\

         \includegraphics[width=0.125\textwidth]{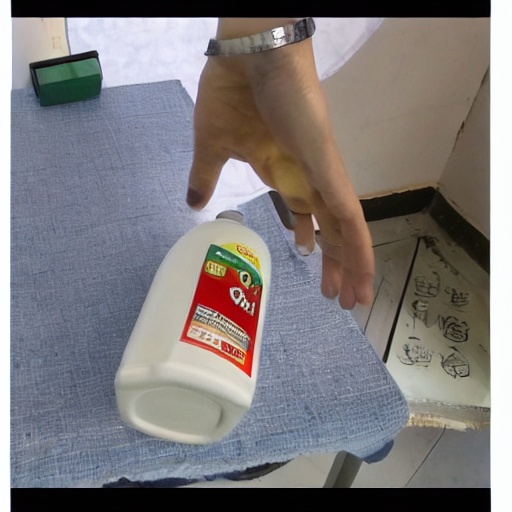} &
         \includegraphics[width=0.125\textwidth]{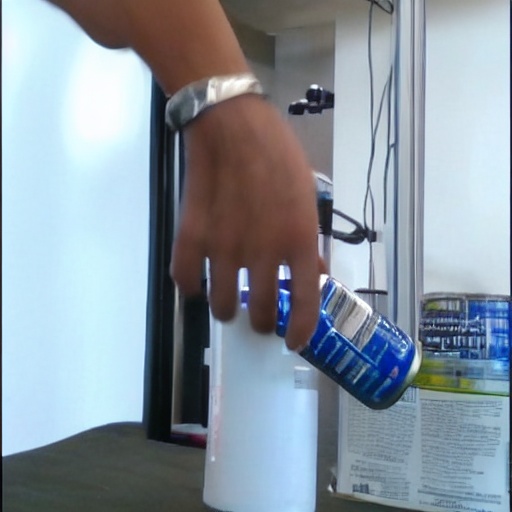}  &
         \includegraphics[width=0.125\textwidth]{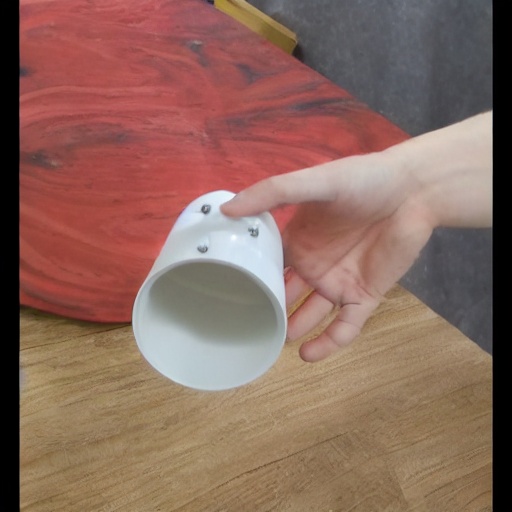}  &
         \includegraphics[width=0.125\textwidth]{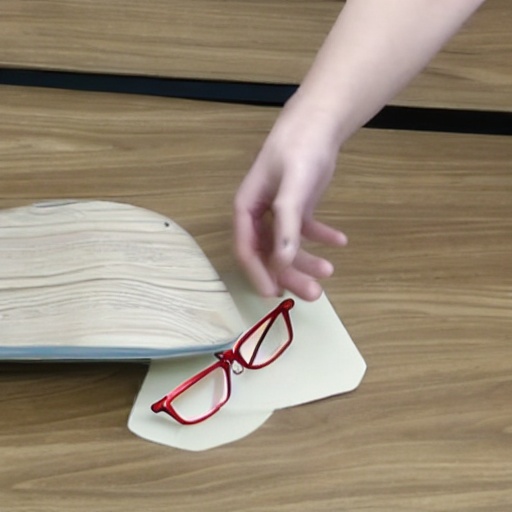} &
         \includegraphics[width=0.125\textwidth]{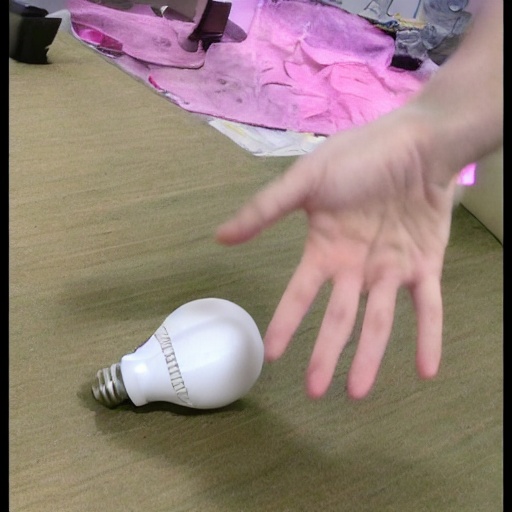}  &
         \includegraphics[width=0.125\textwidth]{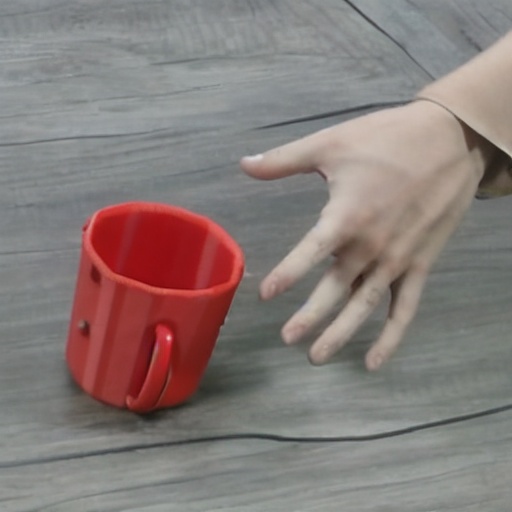} &
         \includegraphics[width=0.125\textwidth]{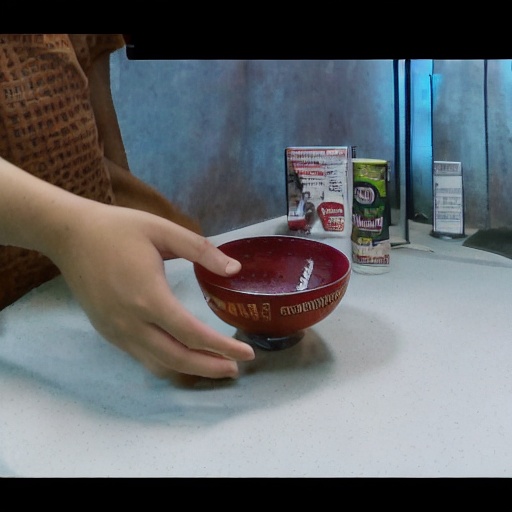} &
         \includegraphics[width=0.125\textwidth]{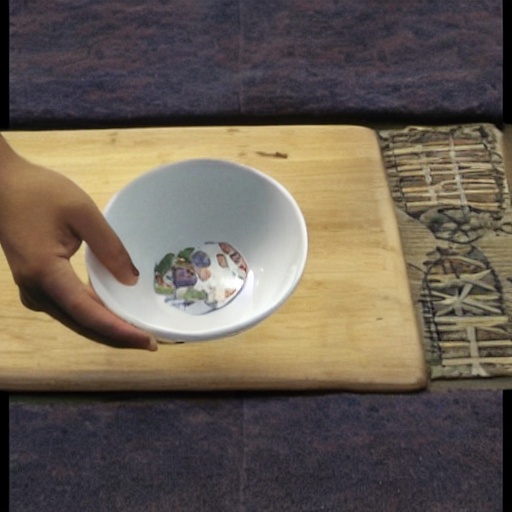} \\
         \\
    \end{tabular}
    }
    \vspace{-3mm}
    \caption{More results on synthesized images with diverse object shapes and hand poses.}
    \vspace{-4mm}
    \label{fig:sup_hoi}
\end{figure*}

The apparent differences in styles between Figure \ref{fig:sup_hoi} and Figure \ref{fig:sup_back} come from finetuning and regularization. In detail, during finetuning, the diffusion model quickly converges to the styles of the training dataset, which is more realistic. Prompts in HOI datasets (in our case, DexYCB~\cite{chao2021dexycb}) lack detailed descriptions of background or in most cases, in a laboratory or the studio environment. Therefore, during inference, if we leave background descriptions vacant or use "on the table", the generated images are much closer to training cases, a bit blurred. To prevent the pretrained model from overfitting, we introduce the regularization module. This allows the model to utilize embedded knowledge from previous scale training, and thus when novel background prompts are provided, our HOIDiffusion is still able to depict diverse images as expected.

\vspace{-2mm}
\section{Social Impact}
\label{sec:social}
During the training process, we use the public well-known hand-object-interaction dataset DexYCB to supervise our training, which is licensed under CC BY-NC 4.0. Our proposed method generates human hands in images. The attributes of generated hands come from the learning knowledge from DexYCB and the pretrained model using LAION~\cite{schuhmann2021laion}, which can be viewed as accumulated average results. Hence, Our synthesized HOI images don't incorporate any personal information or privacy.

\end{document}